\documentclass[runningheads]{llncs}

\usepackage{eccv}

\usepackage{eccvabbrv}

\usepackage{graphicx}
\usepackage{booktabs}

\usepackage{amsmath}
\usepackage{amssymb}
\usepackage{siunitx}
\usepackage{xcolor}
\usepackage{mathtools}
\usepackage{listings}
\usepackage{bm}
\usepackage{enumerate}
\usepackage{pifont}
\usepackage{threeparttable}
\usepackage{float}

\usepackage{multirow}
\usepackage{wrapfig}

\newcommand*\halfcirc[1][1ex]{%
  \begin{tikzpicture}
  \draw[fill] (0,0)-- (90:#1) arc (90:270:#1) -- cycle ;
  \draw[thick] (0,0) circle (#1);
  \end{tikzpicture}}

\usepackage[margin=2.5cm]{geometry}
\usepackage{pgfplots, pgfplotstable}
\pgfplotsset{compat=1.18}
\usepgfplotslibrary{polar}
\usepackage{tikz}
\usetikzlibrary{shapes,arrows,arrows.meta,positioning,calc}
\usepackage{xcolor}
\definecolor{OIblue}{HTML}{0072B2}
\definecolor{colorCoVLA}{HTML}{009E73}
\definecolor{colorScenes}{HTML}{E69F00}
\definecolor{colorOurs}{HTML}{D55E00}
\definecolor{OIgray}{HTML}{7F7F7F}

\providecommand{\saidhl}[1]{{\setlength{\fboxsep}{1.5pt}\colorbox{OIblue!14}{\strut #1}}}

\providecommand{\figcone}[3][1]{%
  \fill[colorScenes] ({#2-0.09*#1},{#3}) -- ({#2+0.09*#1},{#3}) -- ({#2},{#3+0.24*#1}) -- cycle;}

\providecommand{\figcar}[4][1]{%
  \begin{scope}[shift={(#2,#3)}, rotate=#4, scale=#1]
    \fill[black!65, rounded corners=1.6pt] (-0.36,-0.17) rectangle (0.36,0.17);
    \fill[black!25] (0.10,-0.12) rectangle (0.19,0.12);
  \end{scope}}

\tikzset{
  fig1pic/.style={font=\footnotesize\sffamily, line width=0.6pt},
  fig1panel/.style={draw=black!35, rounded corners=3.5pt, line width=0.5pt},
  fig1box/.style={draw=black!55, rounded corners=2.5pt, line width=0.6pt,
                  inner sep=5pt, align=left, font=\footnotesize\sffamily},
  fig1chip/.style={rounded corners=4.5pt, inner xsep=5pt, inner ysep=2.6pt,
                   font=\scriptsize\sffamily, align=center},
  fig1arrow/.style={-{Stealth[length=2.4mm, width=1.9mm]}, line width=0.7pt,
                    draw=black!60},
  fig1traj/.style={line width=1.3pt, -{Stealth[length=2.8mm, width=2.2mm]}},
  fig1label/.style={font=\small\bfseries\sffamily, anchor=base west, inner sep=0pt},
  fig1title/.style={font=\footnotesize\bfseries\sffamily, align=left, inner sep=0pt},
  fig1note/.style={font=\scriptsize\sffamily, text=black!60, align=left},
}

\definecolor{OIblue}{HTML}{0072B2}
\definecolor{colorCoVLA}{HTML}{009E73}
\definecolor{colorScenes}{HTML}{E69F00}
\definecolor{colorWaymo}{HTML}{0583ff}
\definecolor{colorOurs}{HTML}{D55E00}
\definecolor{OIgray}{HTML}{7F7F7F}     

\usepackage[accsupp]{axessibility}  

\usepackage[most]{tcolorbox}
\tcbuselibrary{theorems,breakable}
\usepackage{CJKutf8}
\usepackage{wrapfig}

\usepackage[most]{tcolorbox}
\tcbuselibrary{theorems,breakable}

\definecolor{colorReasoning}{named}{blue}
\definecolor{colorPrompt}{HTML}{009E73}

\DeclareCaptionType[within=none]{promptbox}[Prompt][List of prompts]
\DeclareCaptionType[within=none]{reasoningsbox}[Reasoning][List of reasonings]

\newtcolorbox{prompt}[4][]{
  breakable,
  colback=colorPrompt!5!white,
  colframe=colorPrompt!75!black,
  fonttitle=\bfseries\small,
  fontupper=\small,
  title={#2},
  after={\captionsetup{type=promptbox}\captionof{promptbox}{#3}\label{#4}\vspace{3mm}},
  #1
}

\newtcolorbox{reasoning}[4][]{
  breakable,
  colback=blue!5!white,
  colframe=blue!75!black,
  fonttitle=\bfseries\small,
  fontupper=\small,
  title={#2},
  after={\captionsetup{type=reasoningsbox}\captionof{reasoningsbox}{#3}\label{#4}\vspace{3mm}},
  #1
}

\newdimen\spiderRadius 
\newdimen\spiderLabel 
\usepackage[pagebackref,breaklinks,colorlinks,citecolor=eccvblue,linkcolor=eccvblue,urlcolor=eccvblue]{hyperref}

\usepackage{orcidlink}

\begin{document}

\title{Reasoning models do not yet follow their reasoning \\in autonomous driving: The KITScenes LongTail Dataset} 

\titlerunning{LongTail Driving Scenarios}

\author{
\textbf{Royden Wagner}\thanks{Joint first authorship. $^{\blacktriangle}$ Corresponding author.}\inst{1,2} \and
\textbf{{\"O}mer~{\c{S}}ahin~Ta{\c{s}}}$^{\star}$\inst{2} \and
\textbf{Jaime Villa}$^{\star}$\inst{3} \and
Felix Hauser\inst{2} \and
Yinzhe Shen\inst{1} \and
Marlon Steiner\inst{1} \and
Dominik Strutz\inst{1} \and
Carlos Fernandez$^{\blacktriangle}$\inst{1} \and
Quentin Delfosse\inst{4} \and
Christoph Weinhuber\inst{5} \and
Christian Kinzig\inst{1} \and
Guillermo S. Gutierrez-Cabello\inst{6} \and
Hendrik Königshof\inst{2} \and
Fabian Immel\inst{1,2} \and
Richard Schwarzkopf\inst{1,2} \and
Nils Alexander Rack\inst{1} \and
Kevin Rösch\inst{1,2} \and
Kaiwen Wang\inst{1} \and
Jan-Hendrik Pauls\inst{1} \and
Martin Lauer\inst{1} \and
Igor Gilitschenski\inst{7} \and
Holger Caesar\inst{8} \and
Christoph Stiller\inst{1,2}
}

\authorrunning{Wagner et al.}

\institute{
\textsuperscript{1} Karlsruhe Institute of Technology (KIT)
\quad
\textsuperscript{2} FZI Research Center for Information Technology
\\
\textsuperscript{3} Carlos III University of Madrid
\quad
\textsuperscript{4} Technical University of Darmstadt
\\
\textsuperscript{5} University of Oxford
\quad
\textsuperscript{6} Technical University of Madrid
\\
\textsuperscript{7} University of Toronto
\quad
\textsuperscript{8} Delft University of Technology
}

\maketitle

\begin{abstract}
Handling rare events is the central open challenge in autonomous driving. 
Reasoning models, which generate explicit chains of reasoning before acting, promise to generalize to such events. 
Here we show that these models frequently do not follow their own reasoning: the actions they state in their reasoning often diverge from the actions they ultimately execute. 
We introduce KITScenes LongTail, a curated dataset of rare driving scenarios to quantify this divergence through a measure of semantic reasoning-action coherence. 
We find that incoherence is widespread across current general-purpose and domain-specific models. 
Strikingly, when reasoned actions and executed actions disagree, the reasoning is usually right: extracting actions from the reasoning trace and executing them through a kinematic model enforces coherence and typically improves  motion planning. 
These results suggest that the reasoning capabilities of current models exceed what their actions reveal, and that coherently acting on stated reasoning is a prerequisite for trustworthy autonomous driving.
Our dataset and evaluation space are available at: \texttt{\href{https://hf.co/datasets/kit-mrt/kitscenes-longtail}{hf.co/datasets/kit-mrt/kitscenes-longtail}}
\end{abstract}


\begin{figure}[t]
    \centering
\providecommand{\xtag}[1]{{\ttfamily\textcolor{black!45}{#1}}}
\begin{tikzpicture}[fig1pic]
  \useasboundingbox (0,-0.15) rectangle (16.5,5.0);

  \draw[fig1panel] (0,1.1) rectangle (7.4,4.95);
  \fill[OIblue, rounded corners=1pt] (0.4,4.29) rectangle (0.5,4.65);
  \node[fig1title, anchor=base west] at (0.64,4.37) {What the model \textcolor{OIblue}{reasons}};
  \node[anchor=north west, align=left, text width=6.45cm,
        font=\scriptsize\sffamily, text=black!75] at (0.42,4.15)
    {\xtag{<situational\_awareness>}I am approaching a construction zone;
     the right side of my lane is closed
     ahead.\xtag{</situational\_awareness>}
     \xtag{<acceleration\_first\_3s>}\saidhl{decelerating slightly}
     \xtag{</acceleration\_first\_3s>}
     \xtag{<steering\_first\_3s>}\saidhl{turning slightly left}
     \xtag{</steering\_first\_3s>} \xtag{\ldots}};
  \node[fig1note, anchor=west] at (0.42,1.6) {reasoned:};
  \node[fig1chip, fill=OIblue!12, text=OIblue!40!black, anchor=west] at (1.75,1.6) {decelerate slightly};
  \node[fig1chip, fill=OIblue!12, text=OIblue!40!black, anchor=west] at (4.5,1.6) {steer slightly left};

  \draw[line width=0.8pt, black!45, fill=white] (8.25,3.1) circle (0.52);
  \node[font=\large\boldmath, text=black] at (8.25,3.1) {$\neq$};
  \node[fig1note, align=center, anchor=north] at (8.25,2.38)
    {incoherent\\in 50--97\% \\ of scenarios};

  \draw[fig1panel] (9.1,1.1) rectangle (16.5,4.95);
  \fill[colorOurs, rounded corners=1pt] (9.5,4.29) rectangle (9.6,4.65);
  \node[fig1title, anchor=base west] at (9.74,4.37) {What the model \textcolor{colorOurs}{executes}};

  \fill[black!10] (9.45,2.0) rectangle (16.15,3.9);
  \draw[black!30, line width=0.5pt] (9.45,2.0) -- (16.15,2.0);
  \draw[black!30, line width=0.5pt] (9.45,3.9) -- (16.15,3.9);
  \draw[black!35, dashed, line width=0.5pt] (9.45,2.95) -- (16.15,2.95);
  \fill[colorScenes!14] (12.9,2.0) -- (16.15,2.0) -- (16.15,2.79) -- (14.2,2.79) -- cycle;
  \figcar[1.05]{10.1}{2.48}{0}
  \node[font=\scriptsize\sffamily, text=black!50] at (10.1,2.18) {ego};
  \draw[fig1traj, colorOurs] (10.6,2.48) -- (15.3,2.48);
  \foreach \x in {11.5,12.3,13.1,13.9}{
    \draw[colorOurs!60, line width=0.6pt] (\x,2.37) -- (\x,2.59);}
  \node[font=\scriptsize\sffamily, text=colorOurs!75!black, anchor=east] at (16.05,2.2) {executed};
  \draw[fig1traj, OIblue, dashed]
    (10.6,2.48) .. controls (11.6,2.48) and (12.15,2.55) .. (13.0,2.99)
                .. controls (13.5,3.27) and (13.8,3.35) .. (14.15,3.39);
  \node[font=\scriptsize\sffamily, text=OIblue, anchor=east] at (16.05,3.39) {reasoned};
  \figcone[0.85]{13.05}{2.06} \figcone[0.85]{13.5}{2.3} \figcone[0.85]{13.95}{2.55}
  \figcone[0.85]{14.35}{2.72} \figcone[0.85]{14.95}{2.74} \figcone[0.85]{15.55}{2.74}

  \node[fig1note, anchor=west] at (9.5,1.6) {executed:};
  \node[fig1chip, fill=colorOurs!12, text=colorOurs!40!black, anchor=west] at (10.85,1.6) {maintain speed};
  \node[fig1chip, fill=colorOurs!12, text=colorOurs!40!black, anchor=west] at (13.15,1.6) {steer straight};

\end{tikzpicture}
    \vspace{-1.5cm}
    \caption{\textbf{When reasoned actions and executed actions disagree, the reasoning is usually right.} The actions described in the reasoning trace (left) provide a \textit{better} answer to the long-tail scenario than the executed actions (right). Therefore, executing them with a kinematic model improves motion planning.}
    \label{fig:reasoned_actions_vs_executed_actions}
\end{figure}

\section{Main}

\textbf{Long-tail decision-making, not perception, has long been the bottleneck in autonomous driving and reasoning models have been introduced as the proposed remedy.}
Self-driving has seen substantial progress over the past decade.
Perception, once the primary bottleneck, has advanced significantly through public datasets and benchmarks~\cite{geiger2012kitti, caesar2020nuscenes, sun2020scalability, wilson2023argoverse}.
Today, self-driving cars are deployed across diverse geographical regions (e.g., Waymo), and perception-level generalization has seen significant improvements \cite{madan2024revisiting, xia2025openad}.
However, generalization in perception alone is not sufficient; decision-making in long-tail scenarios remains a major challenge.
In parallel, advances in large language models (LLMs) \cite{openai2024o1, guo2025deepseek} enable contextual generalization and human-interpretable reasoning \cite{ke2025a}, with language serving as a natural medium for expressing goals, constraints, and rationales.
In robotics, embodied chain-of-thought reasoning \cite{zawalski2024robotic, pi2025, team2025gemini}, in which policies verbalize sub-goals and scene-grounded plans before acting, has been shown to improve generalization to unseen tasks and environments.
Building on these advances, reasoning-capable driving models \cite{zhou2025autovla, wang2025alpamayo, nvidia2026alpamayo2, lu2026xiaomi} generate explicit chains of reasoning before acting, promising to generalize to rare events by verbalizing what they perceive and which steering and acceleration actions they intend to execute, and why.

\textbf{Existing evaluations reward either good actions or plausible-sounding reasoning, but no benchmark verifies if the reasoned actions and the executed actions agree.}
Current evaluations of driving models fall into two disconnected camps. 
Many related works on end-to-end motion planning score planned trajectories, typically via $L_2$ errors against expert trajectories \cite{hu2023planning, sun2025sparsedrive, jiang2023vad} or simulation-based composite metrics \cite{jia2024bench2drive, gerstenecker2026fail2drive, dauner2024navsim, cao2025pseudo}. 
Language-centric benchmarks \cite{sima2024drivelm, marcu2024lingoqa, qian2024nuscenes}, in turn, evaluate whether models understand driving scenes through visual question answering, rewarding explanations that sound plausible. 
Neither connects the two: a model can articulate a perfectly reasonable plan like "the right side of my lane is closed ahead, thus I will decelerate and steer slightly left", and then execute a trajectory that maintains speed and steers straight (see  \cref{fig:reasoned_actions_vs_executed_actions}), without any benchmark registering the contradiction. 
If models do not act on their verbalized reasoning, even valid reasons become irrelevant, and the reasoning trace loses its value as an interpretable account of the model's behavior \cite{korbak2025chain}.

\textbf{We introduce a semantic reasoning-action coherence metric (SRAC) to expose that incoherence is common across reasoning models.}
We define semantic reasoning-action coherence (SRAC) as how well the driving actions described in a reasoning trace match the actions in the planned future trajectory: a light-weight, embedding-based Rocchio classifier \cite{manning2008introduction} extracts the steering and acceleration actions expressed in the reasoning and compares them to the actions derived from the planned trajectory (see \cref{subsec:semantic_coherence_metric}). 
Applying this measure to our benchmark of curated long-tail driving scenarios (introduced below), we evaluate driving-specific \cite{zhou2025autovla, lu2026xiaomi, wang2025alpamayo, nvidia2026alpamayo2}, embodied \cite{nvidia2025cosmosreason2, team2025gemini, fang2026molmoact2}, and general-purpose reasoning models \cite{singh2025openai, google2025gemini36flash, anthropic2025claudehaiku45}. 
We find incoherence to be widespread: depending on the model, reasoned and executed actions disagree in 50 to 97\% of scenarios.
This pattern persists when models reason in Spanish or Chinese.

\textbf{To localize the source of incoherence, we first ask whether the reasoned actions are out of place.}
We systematically analyze reasoning-action incoherence and evaluate whether (1) the reasoned actions are worse than the executed actions, so the models rightfully do not execute them, or (2) the reasoned actions are superior and the models are wrong to ignore them.
Notably, we show that the actions expressed in reasoning traces outperform the actions the models actually execute.
Specifically, we parse actions in reasoning traces and execute them through a kinematic model for comparison.

\textbf{We then evaluate how the verbalized reasoning impacts the decoded actions through two interventions.}
First, we remove reasoning entirely by masking thinking tokens from the trajectory decoder or truncating their generation, measuring whether reasoning benefits action decoding in our setting.
Second, we corrupt reasoning traces by raising the sampling temperature until they become nonsensical, measuring whether motion planning is sensitive to the semantic integrity of the reasoning.
We find that planning is only weakly sensitive to corrupted reasoning, and that removing reasoning can improve or degrade motion planning.
Together, these results indicate that executed actions only weakly depend on verbalized reasoning, and that the strength of this coupling is specific to models. 

\textbf{Reasoning and reasoning-action coherence are most relevant in unusual situations. We thus introduce KITScenes LongTail, a curated dataset of uncommon driving scenarios.} 
In nominal driving, there is little to reason about (e.g., maintaining a lane at constant speed requires no explanation), and reasoning traces in such scenarios are trivially coherent with almost any simple trajectory. 
Reasoning is only required in unusual and rare scenarios, where a model must explain how it reacts to circumstances that diverge from nominal driving. Established datasets are dominated by exactly these nominal cases: in nuScenes \cite{caesar2020nuscenes}, driving straight and regular turns account for approx. 90\% of maneuvers. 
We therefore present KITScenes LongTail, one thousand scenarios recorded over two years in urban, suburban, and highway environments, curated around the long tail: road closures, accidents, construction zones, nighttime, adverse weather, and overtaking at high speeds ($>100\, \si{\kilo\meter/\hour}$). 
Each scenario provides six synchronized camera views and is annotated with fine-grained textual instructions (e.g., "overtake truck driving on the right") that enable a precise evaluation of instruction following and context-aware decision-making.
Additionally, we provide reference trajectories for multiple possible maneuvers. 
This focus makes disagreement between reasoning and action both observable and consequential.

In summary, our primary contributions are:
\begin{enumerate}[(i)]
\item We introduce an efficient evaluation of reasoning-action coherence and multiple possible maneuvers. 
\cref{subsec:efficient_eval} covers that our Rocchio classifier outperforms well-established similarity and computationally heavy LLM-as-a-judge metrics.
\cref{fig:ds_mms} shows that our pseudo-simulation approach using the multi-maneuver score (MMS) is a good approximation of closed-loop simulation.

\item We confirm that reasoning models generalize better to our long-tail driving scenarios than non-reasoning-capable end-to-end models.
\cref{tab:reasoning_models_generalize_better} compares standard end-to-end and driving world models to reasoning-capable driving models.
\item We find that despite improved generalization, reasoning models often do not follow their reasoning. Specifically, actions expressed in reasoning traces and executed actions frequently disagree, as demonstrated by the low full coherence rate between the two in \cref{subsec:reasoning_and_action_frequently_disagree}.
\item We reveal that in disagreement between reasoning and predicted actions, the reasoning is usually right. When decoding actions expressed in reasoning traces using a kinematic model, the motion planning performance typically improves. \cref{tab:reasoning_is_usually_right} shows high win rates of this hybrid approach compared to the actions originally executed by the reasoning models.
\item Through reasoning-removal and reasoning-corruption interventions, we demonstrate that motion planning only weakly depends on reasoning traces (see \cref{subsec:executed_actions_only_weakly_depend_on_reasoning}).
\end{enumerate}

\section{Results}
\subsection{Efficiently evaluating reasoning-action coherence and multiple maneuvers}
\label{subsec:efficient_eval}
\newcommand{\stdstack}[2]{\shortstack{#1\\[-2pt]{\fontsize{5}{6}\selectfont\textcolor{gray}{$\pm$#2}}}}

\newcommand{\std}[1]{\,{\fontsize{6}{8}\selectfont\textcolor{gray}{$\pm$#1}}}
\begin{wraptable}{r}{0.55\textwidth}
\vspace{-0.8cm}
    \centering
    {\fontsize{7}{9}\selectfont
    \begin{tabular}{llccc}
        \toprule
        Lang. & Metric & Acceleration & Steering & Avg \\
        \midrule
        \multirow{4}{*}{English}
          & BLEU \cite{papineni-etal-2002-bleu}        & 0.48\std{0.38} & 0.80\std{0.17} & 0.64\std{0.33} \\
          & ROUGE-L \cite{lin-2004-rouge}               & 0.48\std{0.38} & 1.00\std{0.00} & 0.74\std{0.37} \\
          & LingoJudge \cite{marcu2024lingoqa}        & 1.00\std{0.00} & 0.95\std{0.07} & \underline{0.98}\std{0.05} \\
          & RocchioEmb (ours)                          & 0.98\std{0.06} & 1.00\std{0.00} & \textbf{0.99}\std{0.04} \\
        \midrule
        \multirow{4}{*}{Spanish}
          & BLEU \cite{papineni-etal-2002-bleu}        & 0.38\std{0.29} & 0.90\std{0.14} & 0.64\std{0.35} \\
          & ROUGE-L \cite{lin-2004-rouge}               & 0.40\std{0.35} & 0.90\std{0.14} & 0.65\std{0.36} \\
          & LingoJudge \cite{marcu2024lingoqa}        & 1.00\std{0.00} & 0.98\std{0.06} & \textbf{0.99}\std{0.04} \\
          & RocchioEmb (ours)                          & 0.93\std{0.11} & 1.00\std{0.00} & \underline{0.96}\std{0.08} \\
        \midrule
        \multirow{4}{*}{Chinese}
          & BLEU \cite{papineni-etal-2002-bleu}        & 0.55\std{0.38} & 0.88\std{0.13} & 0.71\std{0.32} \\
          & ROUGE-L \cite{lin-2004-rouge}               & 0.55\std{0.38} & 0.90\std{0.14} & 0.73\std{0.33} \\
          & LingoJudge \cite{marcu2024lingoqa}        & 0.80\std{0.17} & 0.88\std{0.13} & \underline{0.84}\std{0.14} \\
          & RocchioEmb (ours)                          & 0.90\std{0.16} & 0.90\std{0.10} & \textbf{0.90}\std{0.13} \\
        \midrule
        \multirow{4}{*}{Overall}
          & BLEU \cite{papineni-etal-2002-bleu}        & \multicolumn{2}{c}{--} & 0.66\std{0.33} \\
          & ROUGE-L \cite{lin-2004-rouge}               & \multicolumn{2}{c}{--} & 0.70\std{0.35} \\
          & LingoJudge \cite{marcu2024lingoqa}        & \multicolumn{2}{c}{--} & \underline{0.93}\std{0.08} \\
          & RocchioEmb (ours)                          & \multicolumn{2}{c}{--} & \textbf{0.95}\std{0.08} \\
        \bottomrule
    \end{tabular}
    }
    \caption{
    \textbf{Synonym robustness: Our light-weight Rocchio classifier outperforms BLEU and ROUGE-L and an LLM-as-a-judge metric.} 
    We measure the robustness w.r.t. synonyms as a classification task. Specifically, we classify synonyms and paraphrased versions of the set of high-level actions shown in the left column in \cref{tab:synonyms}.
    RocchioEmb refers to our approach using an EmbeddingGemma model \cite{vera2025embeddinggemma} with 300M parameters.
    Action labels, synonyms, and paraphrased versions  are included in the extended data section.
    Best average values are \textbf{bold}, second best are \underline{underlined}.
    }
    \label{tab:synonym_robustness}
\end{wraptable}

\subsubsection{Our light-weight Rocchio classifier is more robust than well-established similarity and LLM-as-a-judge metrics.}
We propose to measure semantic reasoning-action coherence (SRAC) with an embedding-based Rocchio classifier (see \cref{subsec:semantic_coherence_metric}).
Here, we benchmark our light-weight approach with 300M parameters against LingoJudge \cite{marcu2024lingoqa}, a 7B parameter LLM-judge trained on driving data.
Additionally, we evaluate the established similarity metrics BLEU \cite{papineni-etal-2002-bleu} and ROUGE-L \cite{lin-2004-rouge}.
We evaluated all approaches across the three languages with the most native speakers worldwide: English, Spanish, and Chinese.
Specifically, we compare
(a) synonym/paraphrase robustness per language (i.e., same driving action expressed in varied wording must score coherent, different action must not) and 
(b) multi-lingual generalization across English, Spanish, and Chinese.
\cref{tab:synonym_robustness} shows that our Rocchio classifier is more robust to synonyms and generalizes better across languages than a 20$\times$ larger LLM-judge and well-established similarity metrics.
Specifically, our method achieves the highest average robustness overall (0.95), although the LLM-judge is slightly more robust in Spanish (0.99 vs.\ 0.96).
The LLM-judge, in turn, is considerably less robust in Chinese, where its average robustness drops to 0.84, driven primarily by lower robustness to synonyms of acceleration actions.
Both BLEU and ROUGE-L are considerably less robust to synonyms of acceleration actions than of steering actions across all three languages.
As shown by the high standard deviations in \cref{tab:synonym_robustness}, the robustness of these similarity metrics significantly varies across different actions.

\subsubsection{Our MMS metric that considers multiple possible maneuvers is a computationally efficient approximation of a closed-loop simulation.} 
\begin{figure}[h!]
    \definecolor{myBlue}{HTML}{0072B2}
    \definecolor{myRed}{HTML}{D55E00}
    \centering
    \includegraphics[width=\linewidth]{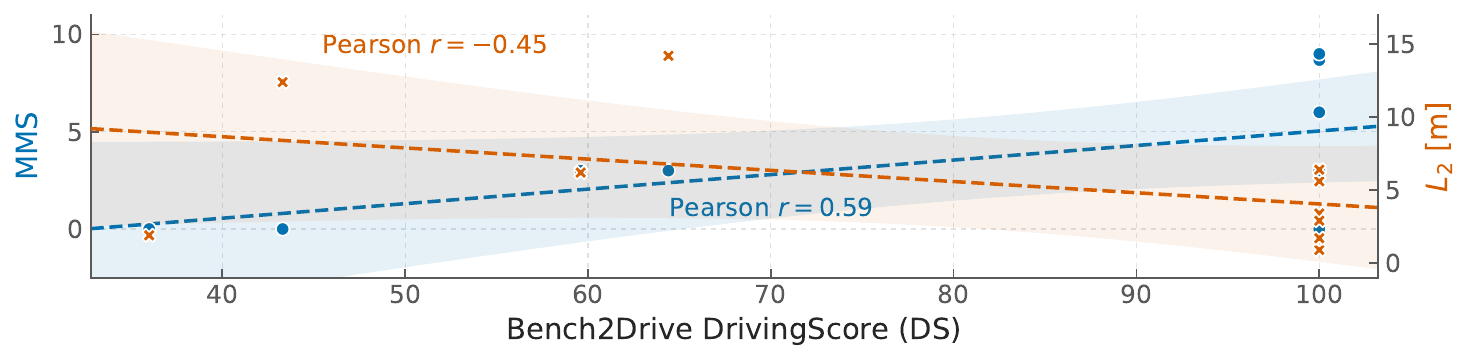}
    \caption{\textbf{Our 
    multi-maneuver scores (MMS) are a closer approximation of closed-loop simulation than L2 errors.
    }
    Showing the relationship between \textcolor{myBlue}{MMS} and \textcolor{myRed}{$L_2$} vs.\ closed-loop DrivingScore (DS) \cite{jia2024bench2drive}, with linear fits and Pearson $r$ values ($0.59$ and $-0.45$).}

    \label{fig:ds_mms}
\end{figure}

Our multi-maneuver score (MMS) evaluates motion planning in autonomous driving (see \cref{subsec:multi_maneuver_score}). 
We quantify the relation of our metric to closed-loop DrivingScores \cite{jia2024bench2drive} and naive $L_2$ errors, which are still commonly used to evaluate end-to-end driving \cite{hu2023planning, hwang2025emma, sun2025sparsedrive}.
To leverage a simulation environment for this comparison, we recorded Bench2Drive \cite{jia2024bench2drive} scenarios using SimLingo \cite{renz2025simlingo} and determined key frames that matched our scenario classes, and labeled reference trajectories for expert driving, crashes, etc. (see \cref{tab:mms}).
We report the MMS values for the future SimLingo trajectories that were held back and average scores for longer Bench2Drive scenarios with multiple key frames.
\cref{fig:ds_mms} shows that MMS values are significantly more correlated with the DrivingScore (DS) metric than $L_2$ errors are (i.e., Pearson $r$ value of $0.59$ vs.\ $-0.45$).
There are few scenarios with a 0 MMS value and a 100 DS value because DS does not measure consistency with past trajectories (see first case in \cref{eq:mms_plan}).
Thus, our metric correctly returns a low score when the planned future trajectory is not well-adapted to the past trajectory.

\subsection{Reasoning models do generalize better to long-tail scenarios}

\begin{wraptable}{r}{0.5\textwidth}
\vspace{-0.7cm}
    \centering
    {\fontsize{7}{9}\selectfont
     \begin{tabular}{lc>{\color{gray!70}}cc>{\color{gray!70}}c}
    \toprule
    \multirow{2}{*}[-0.13em]{Model}& \multicolumn{2}{c}{MMS $\uparrow$} & \multicolumn{2}{c}{$L_2$ $\downarrow$} \\ \cmidrule(lr){2-3}\cmidrule(lr){4-5}
    & mean & median & mean & median \\ \midrule
    UniAD \cite{hu2023planning} & 3.29 & 3.50 & 10.90 & 6.79 \\
    DMAD \cite{shen2025divide} & 3.52 & 3.50 & 10.04 & 5.51 \\
    Epona \cite{zhang2025epona} & 2.74 & 2.50 & 17.67 & 12.57 \\
    \midrule
    AutoVLA \cite{zhou2025autovla} & 3.13 & 2.50 & 22.12 & 7.92 \\
    OneVL \cite{lu2026xiaomi} & 3.91 & 3.50 & 3.70 & 2.72 \\
    Alpamayo 1.5 \cite{wang2025alpamayo} & 4.69 & 3.50 & 3.30 & 2.67 \\
    Alpamayo 2 \cite{nvidia2026alpamayo2} & 5.06 & 3.50 & 2.51 & 1.91 \\
    \bottomrule
    \end{tabular}
    }
    \caption{
    \textbf{
    Reasoning models (lower block) generalize better to our long-tail data.}
    We evaluate on our test split and the metrics are averaged over our scenario types (see \cref{sec:dataset}).
    In addition, the lower mean-to-median ratio demonstrates that the distribution of our proposed multi-maneuver score (MMS, see \cref{subsec:multi_maneuver_score}) is less skewed by outliers than that of standard $L_2$ errors. 
    }
    \label{tab:reasoning_models_generalize_better}
\end{wraptable}

Reasoning-capable driving models like AutoVLA \cite{zhou2025autovla}, OneVL \cite{lu2026xiaomi}, and Alpamayo \cite{wang2025alpamayo, nvidia2026alpamayo2} promise to generalize to long-tail scenarios.
We perform a zero-shot comparison of these models with standard end-to-end driving models (UniAD \cite{hu2023planning} and DMAD \cite{shen2025divide}) and a driving world model (Epona \cite{zhang2025epona}).
All models are evaluated on our test split using publicly available checkpoints.
\cref{tab:reasoning_models_generalize_better} shows that reasoning-capable models generalize better to our scenarios, achieving both higher MMS values and lower $L_2$ errors.
A minor finding is that overall generalization to our long-tail data is still limited.
Specifically, the mean MMS values are mostly between 3 and 5, representing our neglect instruction and wrong speed categories (see \cref{tab:mms}).
Additionally, the lower mean-to-median ratio demonstrates that the distribution of our proposed multi-maneuver score (MMS) is less skewed by outliers than that of standard $L_2$ errors.

\subsection{However, reasoning and action frequently disagree}
\label{subsec:reasoning_and_action_frequently_disagree}
\subsubsection{Reasoned and executed actions frequently disagree, regardless of model type.}
\cref{tab:reasoning_action_incoherence} reports semantic reasoning-action coherence (SRAC, see \cref{subsec:semantic_coherence_metric}) across three types of reasoning models: driving-specific models \cite{zhou2025autovla, lu2026xiaomi, wang2025alpamayo, nvidia2026alpamayo2}, which natively reason about driving actions, as well as embodied \cite{team2025gemini, fang2026molmoact2, nvidia2025cosmosreason2} and general-purpose reasoning models \cite{google2025gemini36flash, singh2025openai, anthropic2025claudehaiku45}, which we prompt with zero-shot chain-of-thought on top of their reasoning capabilities (see \cref{subsec:zero_shot_inference}).
Incoherence is widespread: the full coherence rate (i.e., scenarios in which all reasoned actions are executed with SRAC = 1.0) ranges from 3\% (Molmo2-ER) to at most 50\% (Cosmos-Reason2). 
In other words, even the most coherent model fails to execute at least one of its reasoned actions in half of all scenarios. 
Notably, domain-specific training does not confer coherence: Alpamayo 2 achieves the best motion planning of all evaluated models (MMS of 5.06, $L_2$ of 2.51) yet is fully coherent in only 14\% of scenarios, whereas general-purpose models reach 30–39\% without any driving-specific training. 
Coherence and trajectory quality thus vary independently across models: a model can plan well while rarely acting on its stated reasoning, and vice versa. 
Furthermore, SRAC top2 scores are consistently close to top1 scores, indicating that the measured disagreements are not borderline classification artifacts but semantically distinct maneuvers (see \cref{sec:discussion}).
\vspace{-2mm}
\begin{table*}[htbp]
  \centering
  {\fontsize{7}{9}\selectfont
  \begin{tabular}{>{\centering\arraybackslash}p{2.3cm}lccccc}
    \toprule
    \multirow{2}{*}{Model type} & \multirow{2}{*}{Model} & Full coherence $\uparrow$  & \multirow{2}{*}{SRAC $\uparrow$} & SRAC $\uparrow$ & \multirow{2}{*}{MMS $\uparrow$} & $L_2 \downarrow$ \\
    & & (\%) & & (top2) & & (\si{\meter}) \\
    \midrule
    \multirow{4}{2.5cm}{\centering Driving-specific reasoning models}
      & AutoVLA \cite{zhou2025autovla}
      & 28.6\std{0.5} & 0.59\std{0.01} & 0.69\std{0.00} & 3.13\std{0.06} & 22.12\std{0.09} \\
      & OneVL \cite{lu2026xiaomi}
      & 39.1\std{4.6} & 0.77\std{0.01} & 0.83\std{0.02} & 3.91\std{0.07} & 3.70\std{0.26} \\
      & Alpamayo 1.5 \cite{wang2025alpamayo}
      & 15.9\std{1.8} & 0.61\std{0.02} & 0.72\std{0.03} & 4.69\std{0.04} & 3.30\std{0.26} \\
      & Alpamayo 2 \cite{nvidia2026alpamayo2}
      & 14.0\std{1.1} & 0.56\std{0.01} & 0.67\std{0.02} & 5.06\std{0.09} & 2.51\std{0.03} \\
    \midrule
    \multirow{3}{2.5cm}{\centering Embodied reasoning models}
      & Cosmos-Reason2 \cite{nvidia2025cosmosreason2}
      & 50.3\std{2.4} & 0.67\std{0.01} & 0.68\std{0.01} & 3.65\std{0.07} & 8.58\std{1.57} \\
      & Gemini Robotics ER 1.6 \cite{team2025gemini}
      & 43.5\std{1.8} & 0.77\std{0.01} & 0.80\std{0.01} & 4.40\std{0.06} & 3.85\std{0.07} \\
      & Molmo2-ER \cite{fang2026molmoact2}
      & 2.7\std{1.5}  & 0.50\std{0.01} & 0.51\std{0.01} & 2.15\std{0.06} & 19.14\std{0.50} \\
    \midrule
    \multirow{3}{2.5cm}{\centering General-purpose reasoning models}
      & GPT-5.6 Luna \cite{singh2025openai}
      & 38.9\std{1.2} & 0.73\std{0.01} & 0.76\std{0.01} & 4.05\std{0.13} & 4.01\std{0.19} \\
      & Gemini 3.6 Flash \cite{google2025gemini36flash}
      & 33.4\std{2.0} & 0.71\std{0.01} & 0.75\std{0.01} & 4.18\std{0.14} & 4.31\std{0.13} \\
      & Claude Haiku 4.5 \cite{anthropic2025claudehaiku45}
      & 29.7\std{1.3} & 0.61\std{0.01} & 0.66\std{0.00} & 3.57\std{0.02} & 8.36\std{0.24} \\
    \bottomrule
  \end{tabular}%
  }
    \caption{
    \textbf{Reasoned and executed actions are frequently incoherent.}
    Full coherence gives the percentage of scenarios where \textit{all} reasoned actions are correctly executed (i.e., where SRAC = 1.0). Semantic reasoning-action coherence (SRAC) gives the average coherence rate across the 4 steering and acceleration actions expressed in reasoning traces (see \cref{subsec:semantic_coherence_metric}).
    For open-weight models, we compute the mean and standard deviation using 3 global seeds. For API-gated models, we call the API 3 times to compute means and standard deviations.
    }
  \label{tab:reasoning_action_incoherence}
\end{table*}

\vspace{-5mm}
\subsubsection{Incoherence persists when models output reasoning in Spanish or Chinese.}
Since reasoning abilities vary across languages in other contexts like math \cite{wang2026polymath}, we test whether the observed incoherence is specific to reasoning in English. 
We repeat our evaluation with the general-purpose reasoning models prompted to reason in Spanish and Chinese.\footnote{
We only perform this multilingual experiment with general-purpose models because embodied and driving-specific reasoning models tend to answer in English, even when instructions are provided in Spanish or Chinese.
}
\cref{tab:multilingual_incoherence} shows that reasoned and executed actions remain frequently incoherent in both languages: fully coherent scenarios range from 21\% to 42\%, and SRAC values remain in a similar range as for English. 
Compared to reasoning in English (see lower block in \cref{tab:reasoning_action_incoherence}), most scores worsen slightly for GPT-5.6 Luna and Claude Haiku 4.5, while Gemini 3.6 Flash appears less affected by the change in language. 
Semantic reasoning-action incoherence is therefore not an artifact of one reasoning language, but a consistent property of current reasoning models.

\begin{table}[htbp]
\centering

{\fontsize{7}{9}\selectfont
\begin{tabular}{llccccc}
\toprule
\multirow{2}{*}{Language} & \multirow{2}{*}{Model} & Full coherence $\uparrow$  & \multirow{2}{*}{SRAC $\uparrow$} & SRAC $\uparrow$ & \multirow{2}{*}{MMS $\uparrow$} & $L_2 \downarrow$ \\
    & & (\%) & & (top2) & & (\si{\meter}) \\
\midrule
\multirow{3}{*}{Spanish}  
& GPT-5.6 Luna \cite{singh2025openai} & 35.3\std{0.7} & 0.71\std{0.01} & 0.75\std{0.01} & 3.98\std{0.10} & 4.53\std{0.32} \\
& Gemini 3.6 Flash \cite{google2025gemini36flash} & 34.1\std{0.7} & 0.70\std{0.01} & 0.74\std{0.01} & 4.13\std{0.09} & 4.62\std{0.29} \\
& Claude Haiku 4.5 \cite{anthropic2025claudehaiku45} & 23.7\std{0.8} & 0.57\std{0.01} & 0.64\std{0.01} & 3.39\std{0.07} & 8.42\std{0.33} \\
\midrule
\multirow{3}{*}{Chinese} 
& GPT-5.6 Luna \cite{singh2025openai} & 35.3\std{1.8} & 0.74\std{0.01} & 0.76\std{0.01} & 3.92\std{0.12} & 4.09\std{0.25} \\
& Gemini 3.6 Flash \cite{google2025gemini36flash} & 42.0\std{1.9} & 0.78\std{0.00} & 0.81\std{0.00} & 4.20\std{0.07} & 4.41\std{0.12} \\
& Claude Haiku 4.5 \cite{anthropic2025claudehaiku45} & 21.0\std{1.2} & 0.60\std{0.02} & 0.72\std{0.01} & 3.26\std{0.05} & 8.60\std{0.21} \\
\bottomrule
\end{tabular}
}
\caption{
\textbf{When reasoning in Spanish or Chinese, reasoned and executed actions are frequently incoherent as well.}
Compared to reasoning in English (see lower block in \cref{tab:reasoning_action_incoherence}), most scores worsen slightly for GPT-5.6 Luna and Claude Haiku 4.5, while Gemini 3.6 Flash seems to be less affected by the change in language.
}
\label{tab:multilingual_incoherence}
\end{table}

\subsection{In disagreement, the reasoning is usually right}
\label{subsec:reasoning_is_usually_right}

\subsubsection{When reasoned and executed actions disagree, the reasoned actions usually score higher.}
Having established that reasoning and action frequently disagree, we next resolve which side of the disagreement is right. 
For incoherent cases, we score both the maneuver expressed in the reasoning trace and the maneuver implied by the decoded trajectory against our possible maneuver references (see \cref{subsec:multi_maneuver_score}), and report the rate at which the reasoned actions win. 
We filter out ties, yielding an adjusted win rate. 
As shown in \cref{tab:reasoning_is_usually_right}, the reasoned actions win in the clear majority of disagreements: win rates reach 86\% for AutoVLA, 85\% for Claude Haiku 4.5, 80\% for OneVL and Molmo2-ER, and 76\% for Cosmos-Reason2. 
The reasoning is usually right and the models discard better actions than the ones they execute.

\subsubsection{Executing reasoned actions through a kinematic model typically improves motion planning and enforces coherence by design.}
Specifically, we parse the actions expressed in reasoning traces and execute them through a kinematic model (see \cref{sec:kinematic_model}). 
This inference setting improves motion planning for 8 of the 10 evaluated models, with average MMS gains of up to 1.34 (Molmo2-ER). 
We measure the largest gains for models whose decoded trajectories are weakest, indicating that their verbalized reasoning contains planning capability that their action decoding fails to realize. 
The two exceptions are Alpamayo 1.5 and Alpamayo 2 (average MMS changes of -0.39 and -0.37), the strongest motion planners in our evaluation, whose decoded trajectories outperform their stated actions. 
In addition to the performance gains, this setting enforces reasoning-action coherence by design, as reflected by SRAC scores of 0.88 and above across all models (see \cref{tab:reasoning_is_usually_right}). 
Together, these results localize the dominant failure mode to action generation rather than reasoning: the models reason adequately about what to do, but do not decode trajectories that follow it.

\begin{table}[htbp]
  \centering
 
  {\fontsize{7}{9}\selectfont
  \begin{tabular}{>{\centering\arraybackslash}p{2.3cm}lccccc}
    \toprule
     \multirow{2}{*}{Model type} &  \multirow{2}{*}{Model} & Win rate $\uparrow$ &  \multirow{2}{*}{MMS $\uparrow$} & $L_2$ $\downarrow$ & \multirow{2}{*}{SRAC $\uparrow$} &  SRAC $\uparrow$ \\
    & & (\%) & & (\si{\meter}) & & (top2) \\
    \midrule
    \multirow{4}{2.5cm}{\centering Driving-specific reasoning models}
    & AutoVLA \cite{zhou2025autovla} (kinematic)          & 86.0\std{1.4} & 3.90\std{0.04} & 8.95\std{0.25} & 0.88\std{0.01} & 0.88\std{0.00} \\
          & OneVL \cite{lu2026xiaomi} (kinematic)  & 80.0\std{2.8} & 4.42\std{0.01} & 3.63\std{0.26} & 1.00\std{0.00} & 1.00\std{0.00} \\
          & Alpamayo 1.5 \cite{wang2025alpamayo} (kinematic)     & 46.2\std{3.0} & 4.30\std{0.07} & 3.45\std{0.05} & 0.98\std{0.01} & 0.98\std{0.01} \\
          & Alpamayo 2 \cite{nvidia2026alpamayo2} (kinematic)       & 37.9\std{1.5} & 4.69\std{0.09} & 3.27\std{0.08} & 0.96\std{0.02} & 0.97\std{0.02} \\
    \midrule
    \multirow{3}{2.5cm}{\centering Embodied reasoning models}
      & Cosmos-Reason2 \cite{nvidia2025cosmosreason2} (kinematic)        & 75.8\std{1.8} & 4.01\std{0.05} & 6.40\std{0.67} & 0.93\std{0.00} & 0.94\std{0.00} \\
          & Gemini Robotics ER 1.6 \cite{team2025gemini} (kinematic) & 67.9\std{4.5} & 4.55\std{0.03} & 3.50\std{0.05} & 0.99\std{0.00} & 0.99\std{0.00} \\
          & Molmo2-ER \cite{fang2026molmoact2} (kinematic)             & 79.8\std{1.7} & 3.49\std{0.04} & 6.58\std{0.06} & 0.99\std{0.00} & 0.99\std{0.00} \\
    \midrule
    \multirow{3}{2.5cm}{\centering General-purpose reasoning models}
      & GPT-5.6 Luna \cite{singh2025openai} (kinematic)     & 60.5\std{2.9} & 4.10\std{0.08} & 3.99\std{0.06} & 0.97\std{0.01} & 0.98\std{0.01} \\
          & Gemini 3.6 Flash \cite{google2025gemini36flash} (kinematic) & 67.6\std{5.1} & 4.31\std{0.12} & 3.64\std{0.10} & 0.98\std{0.00} & 0.99\std{0.00} \\
          & Claude Haiku 4.5 \cite{anthropic2025claudehaiku45} (kinematic) & 85.1\std{0.7} & 3.89\std{0.01} & 6.38\std{0.03} & 0.97\std{0.00} & 0.97\std{0.00} \\
    \midrule
    \multicolumn{2}{l}{Ground truth actions executed through our kinematic model} & - & 5.93 & 1.89& 1.00 & 1.00 \\
    \bottomrule
  \end{tabular}
}
 \caption{
 \textbf{
 When reasoned and executed actions disagree, executing the reasoned actions through a kinematic model typically improves results.
 }
 Win rate measures how often MMS improves when executing reasoned actions through the kinematic model compared to the originally executed actions, after filtering ties (the MMS values of the originally executed actions are shown in \cref{tab:reasoning_action_incoherence}).
 Furthermore, this hybrid inference setting enforces semantic reasoning-action coherence by design, as shown by high SRAC scores.
 As reference for the upper bound, the last row shows the scores obtained by executing the ground truth actions (extracted from the expert trajectory) through our kinematic model (see \cref{sec:kinematic_model}).
 }
  \label{tab:reasoning_is_usually_right}
\end{table}

\subsection{Executed actions only weakly depend on verbalized reasoning}
\label{subsec:executed_actions_only_weakly_depend_on_reasoning}
\subsubsection{Removing reasoning entirely when decoding actions only slightly affects motion planning.}
\label{subsubsec:no_cot_inference}
\cref{subsec:reasoning_is_usually_right} leaves open why the Alpamayo models are the exception in which decoded trajectories outperform the stated actions. 
We hypothesize that models differ in how strongly their action decoding depends on the verbalized reasoning in the first place. 
We therefore remove verbalized reasoning altogether, masking thinking tokens from the trajectory decoder (Alpamayo 1.5, Alpamayo 2) or truncating their generation (AutoVLA, OneVL), depending on API constraints and inference configurations. 
\cref{tab:removing_reasoning} shows that motion planning results slightly improve for Alpamayo 1.5, Alpamayo 2, and AutoVLA when reasoning is removed, and slightly worsen for OneVL.
Overall, the performance changes are rather marginal and no model degrades substantially without its reasoning. Thus, executed actions only weakly depend on the verbalized reasoning, and the strength of this coupling is specific to models.

\begin{table}[htbp]
  \centering
  {\fontsize{7}{9}\selectfont
  \begin{tabular}{lcccc}
        \toprule
        Model & MMS $\uparrow$ & $\Delta$MMS & $L_2$ $\downarrow$ & $\Delta$$L_2$ \\
        \midrule
        Alpamayo 1.5 \cite{wang2025alpamayo}                         & 4.69\std{0.04} &       & 3.30\std{0.26}  &       \\
        Alpamayo 1.5 (masked CoT)             & 4.81\std{0.08} & +0.12 & 2.95\std{0.15}  & $-$0.35 \\
        AutoVLA \cite{zhou2025autovla}                              & 3.13\std{0.06} &       & 22.12\std{0.09} &       \\
        AutoVLA (no CoT)                      & 3.30\std{0.01} & +0.16 & 21.28\std{0.02} & $-$0.84 \\
        OneVL \cite{lu2026xiaomi}                       & 3.91\std{0.07} &       & 3.70\std{0.26}  &       \\
        OneVL (no latent reasoning) & 3.85\std{0.12} & $-$0.06 & 3.75\std{0.09} & +0.05 \\
        Alpamayo 2 \cite{nvidia2026alpamayo2}                            & 5.06\std{0.09} &       & 2.51\std{0.03}  &       \\
        Alpamayo 2 (masked CoT)               & 5.10\std{0.07} & +0.04 & 2.44\std{0.01}  & $-$0.06 \\
        \bottomrule
    \end{tabular}
  }
    \caption{
    \textbf{Removing reasoning only slightly affects executed actions.}
     We remove reasoning by masking thinking tokens from the trajectory decoder (Alpamayo 1.5, Alpamayo 2) or truncating their generation (AutoVLA, OneVL).
     Overall, the changes in MMS and $L_2$ are rather marginal and close to the standard deviations between runs.
    }
  \label{tab:removing_reasoning}
\end{table}

\subsubsection{Corrupting reasoning traces until they are nonsensical only moderately affects executed actions.}
\begin{wrapfigure}{r}{0.5\textwidth}
\vspace{-1cm}
  \centering
  \includegraphics[width=0.9\linewidth]{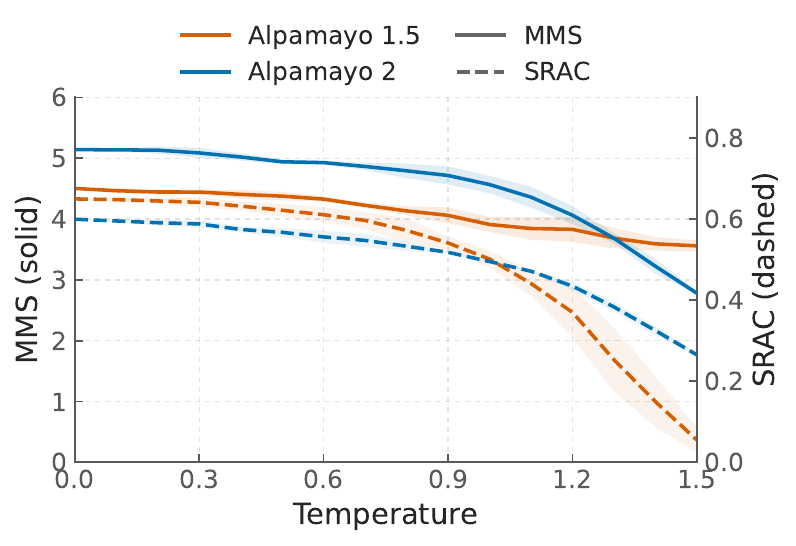}
  \hfill
  \caption{\textbf{Raising sampling temperature until reasoning traces become nonsensical only moderately affects executed actions.}
    Specifically, semantic reasoning-action coherence (SRAC) degrades significantly more with sampling temperature than the multi-maneuver scores (MMS) of executed actions. 
}
    \label{fig:increasing_sampling_temperature}
\end{wrapfigure}

The removal intervention in \cref{tab:removing_reasoning} tests whether trajectory decoding (i.e., the executed actions) benefits from access to reasoning content.
We further test whether the semantic integrity of that reasoning content matters. 
Specifically, we raise the sampling temperature during reasoning generation until the traces become nonsensical, and measure the effect on motion planning. 
We restrict this intervention to Alpamayo 1.5 and Alpamayo 2, where the reasoning temperature can be varied independently of trajectory decoding.\footnote{
For OneVL, the verbalized reasoning is not connected to the predicted actions, as the actions are decoded from latent reasoning tokens generated before the text. Therefore, corrupting the verbalized reasoning would not affect the action decoding. 
For AutoVLA, reasoning and actions are decoded autoregressively under a shared temperature parameter, so any degradation could stem from noisier trajectory decoding rather than from the corrupted reasoning, confounding the intervention. 
} 
\cref{fig:increasing_sampling_temperature} shows that for both Alpamayo models, SRAC degrades significantly more with sampling temperature than MMS. Thus, even when reasoning traces are hallucinated or nonsensical, the models still decode largely sensible trajectories. 
This indicates that trajectory decoding largely bypasses the semantic content of the reasoning trace, likely attending to visual context (e.g., the visual KV cache) rather than to the tokens of the reasoning itself (see discussion in \cref{sec:discussion}). 
We include examples of nonsensical reasoning traces in the extended data section (see \cref{subsec:nonsensical_reasoning_examples}).

\section{Discussion}
\label{sec:discussion}
\subsubsection{Incoherence in current models is mostly a good planner trapped behind a bad decoder.} 
For 8 of the 10 evaluated models, our results localize reasoning-action incoherence to action generation rather than to the reasoning itself. 
The reasoning is not deficient: the reasoned actions are often superior and the models are wrong to ignore them (i.e., the actions expressed in reasoning traces outperform the actions the models actually execute).
Thus, executing reasoned actions through a kinematic model typically improves motion planning compared to the originally executed actions (see \cref{subsec:reasoning_is_usually_right}). 
Our two interventions explain this mismatch since executed actions only weakly depend on verbalized reasoning. 
Incoherence thus arises from a good planner trapped behind a bad decoder: the verbalized reasoning describes sound actions, but the action decoding is only loosely coupled to it and fails to follow them. 
This has three implications. 
First, reasoning traces are a more trustworthy audit surface than commonly assumed: rather than masking the model's decision process, they describe better actions than the ones executed.
Second, hybrid architectures that treat language-level actions as the interface, with classical kinematic and control layers below, are a principled design rather than an engineering workaround.
Our coherence-by-construction setting (see \cref{sec:kinematic_model}) is a minimal instance of such an architecture. 
Third, enforcing coherence by construction shifts the safety question from "does the model mean what it says" to "is what it says any good", and the latter is directly measurable with motion planning metrics.
Our findings thereby differ from research on chain-of-thought faithfulness in general-purpose settings \cite{lanham2023measuring, turpin2023language, palod2025performative, chen2025reasoning}, which shows that reasoning traces can misrepresent the process behind an answer.
Our results suggest that the stated reasoning is not a misleading account of a better hidden process, but a better plan that the model fails to execute. 
However, our findings align with recent observations in robotics that current reasoning models get the most traction at a higher level of abstraction \cite{berman2026clauderobotics} (i.e., a language model acts as a high-level planner while a pretrained low-level policy handles the physics). Language-level actions decoded through a kinematic model follow exactly this structure.

\subsubsection{For the Alpamayo models, the contrary holds: a good decoder ignoring a worse planner.} 
Alpamayo 1.5 and Alpamayo 2 are the strongest motion planners in our evaluation, yet they are among the least coherent (full coherence of 16\% and 14\% in \cref{tab:reasoning_action_incoherence}). Unlike the other models, however, their decoded trajectories outperform their stated actions: executing the reasoned actions through the kinematic model lowers MMS by 0.39 and 0.37, and the reasoned actions win only 46\% and 38\% of disagreements (see \cref{tab:reasoning_is_usually_right}). 
This is the case that most resembles the unfaithful chain-of-thought reported in general-purpose settings \cite{lanham2023measuring, turpin2023language, chen2025reasoning}.
Alpamayo models thus also do not follow their reasoning, but arguably rightfully so.

\subsubsection{Information flow around verbalized reasoning.}
Our interventions show that action decoding is not significantly correlated with the semantic quality of the reasoning: removing reasoning entirely leaves motion planning largely intact or even slightly improves it, and corrupting reasoning traces until they are nonsensical only moderately affects the executed actions (see \cref{subsec:executed_actions_only_weakly_depend_on_reasoning}).
Information therefore likely flows "around" the verbalized reasoning. 
For Alpamayo 1.5 and Alpamayo 2, this is possible through attention to the visual KV cache rather than to the tokens of the reasoning trace.
For AutoVLA, attention between action and reasoning tokens can be limited.
OneVL decodes actions from latent reasoning tokens and the verbalized reasoning is generated by an auxiliary decoder.
These results are related to the phenomenon of improved no-CoT task completion \cite{gould2026think}, where models perform sufficiently complex reasoning internally without emitting thinking tokens or stating explicit reasoning steps.

\subsubsection{Coherently acting on stated reasoning is a prerequisite for trustworthy autonomy.}
Related work \cite{marcu2024lingoqa, sima2024drivelm, qian2024nuscenes} evaluates whether models understand driving scenarios and road users, directly probing visual reasoning via visual question answering (VQA). 
In contrast to our work, this does not connect reasoning to acting: a model can answer scene-understanding questions correctly and still execute actions that contradict its stated reasoning. 
If models do not act on their verbalized reasoning, even valid reasons become irrelevant. 
High semantic reasoning-action coherence is therefore a prerequisite for evaluating, and ultimately trusting, reasoning models in autonomous driving.

\subsubsection{Most common failure modes beyond incoherence.}
First, models do not follow driving instructions precisely: the mean MMS of 4 across most reasoning models corresponds to our \textit{neglect instruction} category (see \cref{tab:mms}), meaning the typical planned trajectory is safe but ignores the high-level instruction. 
Second, models copy the past trajectory into the output as the future trajectory. 
This is common for the smaller embodied reasoning models (Molmo2-ER and Cosmos-Reason2) and is covered by our fallback logic (see \cref{subsec:multi_maneuver_score}), because such trajectories are not executable by downstream motion controllers. 
This behavior relates to the general copy-paste tendency of LLMs in contextual question answering \cite{phukan2024peering}.

\subsection{Limitations and future work}

\subsubsection{Our motion planning evaluation is open-loop and non-reactive.}
The MMS metric evaluates planned trajectories without simulating the reactions of other traffic participants. 
We mitigate this in two ways: (1) our scenarios focus on cases where the ego vehicle has to react to its environment rather than the reverse, and (2) our multi-maneuver reference trajectories cover multiple admissible reactions instead of a single expert trajectory. 
Empirically, MMS correlates well with closed-loop DrivingScores (Pearson r of 0.59, see \cref{fig:ds_mms}), indicating that this open-loop approximation is informative for the scenarios we consider. 
Reactive or neurally rendered closed-loop evaluation of reasoning-action coherence remains future work.

\subsubsection{Parsing as a possible confound.}
A weak parser can lead to the reasoning-action incoherence we report: if actions expressed in reasoning traces were systematically misread, disagreement would be an artifact of measurement rather than of the models. 
We preempt this with the parsing-robustness ablation in \cref{subsec:efficient_eval}, where our Rocchio classifier is robust to synonyms and paraphrases across English, Spanish, and Chinese, and outperforms a driving-specific LLM judge. 
A related question is whether models fail to stick to our output format rather than fail to reason. 
Two observations argue against this: SRAC top2 results are close to the top1 results, so slight mismatches in the semantic mapping are uncommon, and non-parseable outputs are handled explicitly by our fallback logic rather than being scored as incoherent maneuvers.

\subsubsection{Verbalized reasoning might differ from the models' actual reasoning.}
When a model is trained or prompted to state its reasoning and then output a trajectory, hidden reasoning still happens upstream in latent space, and what we score is a verbalized rationale written in the final answer channel. 
Our claims therefore mostly concern the coherence between stated reasoning and executed actions, not the model's internal decision process. 
Analyzing latent-space reasoning \cite{zhu2025survey} and the causal impact of specific reasoning tokens or sentence-pieces on the output (e.g., via thought anchors \cite{bogdan2025thought}) are directions for future work that would connect semantic coherence to mechanistic accounts of how these models decide to act.

\section{Methods}
\subsection{Semantic reasoning-action coherence metric}
\label{subsec:semantic_coherence_metric}
We use Rocchio classification \cite{manning2008introduction} and sentence embeddings to measure semantic coherence between reasoning traces and planned trajectories. 
We define semantic reasoning-action coherence (SRAC) as how well the driving actions described in the reasoning traces match the actions in the planned future trajectory. 
Following \cite{tas2025word}, we apply heuristics to classify the driving actions (i.e., steering and acceleration commands) of a given planned trajectory (see \cref{subsec:heuristics_action_classifying}). 
Then, we generate embeddings of the corresponding segment of reasoning traces using EmbeddingGemma 0.3B~\cite{vera2025embeddinggemma}.
Afterwards, we perform Rocchio classification on these embeddings, comparing them to reference embeddings that represent all possible driving actions according to our taxonomy. 

\begin{equation} \nonumber
    \hat{y} = \arg \max_{c \in \bm{C}} \cos\bigl( \bm{z}, \boldsymbol{\mu}_c \bigr)
\end{equation} 
where $\bm{C}$ is the set of all classes, $\bm{z}$ is an embedding, $\boldsymbol{\mu}_c$ is the reference embedding of class $c$, and $\cos(\cdot)$ computes the cosine similarity.

Finally, we calculate the SRAC metric, which is the rate with which the driving action predicted from the reasoning traces $\hat{y}$ matches the driving action derived from the planned trajectory.
The classification accuracy thus indicates whether the driving actions described in the reasoning traces semantically align with those in the final planned trajectory, quantifying semantic coherence.\footnote{Our approach is related to recent methods for reward generation when training general purpose reasoning models \cite{li2025reinforcement}. Conceptually, low semantic reasoning-action coherence is also related to low CoT faithfulness \cite{lanham2023measuring, turpin2023language, chen2025reasoning}. 
Specifically, low coherence in model outputs suggests that the CoT does not accurately describe the process that led to its actions.}

\subsection{Multi-maneuver scores for motion planning}
\label{subsec:multi_maneuver_score}
\subsubsection{Scoring deviation from only one expert trajectory ignores the fact that multiple maneuvers may be appropriate in a scenario and simulation is expensive.}
We agree with the recent criticism that $L_2$ errors with respect to expert trajectories do not capture the multi-modality of driving \cite{dauner2024navsim, jia2024bench2drive, caesar2021nuplan}.
Specifically, these evaluations overlook the fact that, in many scenarios, multiple maneuvers are appropriate.
Related benchmarks \cite{jia2024bench2drive, gerstenecker2026fail2drive, gorgulu2026tacarla} leverage the CARLA simulator \cite{dosovitskiy2017carla} to run reactive, closed-loop simulation.
However, sensor data simulated in CARLA exhibits a large sim-to-real domain gap.
Furthermore, neural rendering~\cite{ljungbergh2024neuroncap, agarwal2025cosmos, gao2024vista, mousakhan2025orbis, yang2025drivearena} for realistic sensor simulation is promising, yet computationally expensive and prone to visual artifacts.

\begin{wraptable}{r}{0.5\textwidth}
    \vspace{-0.75cm}
    \centering
    \normalsize
    
  {\fontsize{7}{9}\selectfont
    \begin{tabular}{llS}
        \toprule
        Category & Comfort penalty (CP) & {Score} \\
        \midrule
        \multirow{3}{*}{Expert-like trajectory} & none & 10 \\
         & jerk \texttt{XOR} tortuosity & 9 \\
         & jerk \texttt{AND} tortuosity & 8 \\
         \midrule
         \multirow{3}{*}{Wrong speed} & none & 7 \\
         & jerk \texttt{XOR} tortuosity & 6 \\
         & jerk \texttt{AND} tortuosity & 5 \\
         \midrule
         \multirow{3}{*}{Neglect instruction} & none & 4 \\
         & jerk \texttt{XOR} tortuosity & 3 \\
         & jerk \texttt{AND} tortuosity & 2 \\
         \midrule
         Driving off road w/o crashing & not considered & 1 \\
         \midrule
         Crash & not considered & 0 \\
        \bottomrule
    \end{tabular}}
     \caption{\textbf{Reference multi-maneuver scores (MMS).} Our metric ranks planned trajectories based on similarity to reference trajectories of 5 categories. For the first three categories, we apply comfort penalties if the jerk or tortuosity significantly exceeds that of the reference trajectory.}
    \label{tab:mms}
    \vspace{-0.5cm}
\end{wraptable}

\subsubsection{We propose a pseudo-simulation approach to cover multiple possible maneuvers.}
Our multi-maneuver score (MMS) ranks planned trajectories based on similarity to reference trajectories\footnote{Our metric is related to rater-feedback scores~\cite{waymo2025e2e}, but explicitly considers instruction following, comfort, and crashes.} and comfort.
For each scenario, we provide 3 reference trajectories according to the categories in \cref{tab:mms}.
For the \textit{expert-like trajectory} category, we use the trajectory driven by an expert.
For the \textit{wrong speed} category, we augment the expert trajectory using state estimation and spline modifications.
Specifically, we use an extended Kalman filter to smooth the expert trajectory and spline modifications to change the average speed by \SI{\pm 20}{\percent}.
For the \textit{neglect instruction} category, we manually annotate reasonable trajectories that do not follow the high-level instruction. 
For instance, at an intersection where the instruction is to turn right, we provide a trajectory for turning left or driving straight.
For the \textit{driving off road w/o crashing} category, we manually label trajectories in which the ego-vehicle partially or completely leaves the drivable area.
In the \textit{crash} category, we manually label rear-end collisions and crashes involving static obstacles, such as traffic signs or buildings.
We cover comfort by subtracting a comfort penalty $\text{CP} \in \{0, 1, 2\}$ from the maximum MMS value of each category.
Specifically, we consider jerk and tortuosity relative to reference trajectories. 
We compute jerk using
\begin{equation} \nonumber
    \text{average jerk} = \frac{1}{T - 3} \sum_t \left\Vert \frac{\Delta^3 \bm{Y}_{t,:}}{\Delta t^3} \right\Vert,
\end{equation}
where $\bm{Y} \in \mathbb{R}^{T \times 2}$ is a trajectory as a temporal sequence of waypoints with x- and y-coordinates and $t \in \{1,\dots, T\}$ indexes the temporal dimension.
Moreover, we compute tortuosity using
\begin{equation} \nonumber
    \text{tortuosity} = \frac{\sum_{t=2}^{T} \left\Vert \bm{Y}_{t,:} - \bm{Y}_{t-1,:} \right\Vert }{\left\Vert \bm{Y}_{T,:} - \bm{Y}_{1,:} \right\Vert}.
\end{equation}
We reduce the MMS value by 1 if the jerk of a planned trajectory is more than \SI{44}{\percent} higher than that of a reference trajectory.
Similarly, we reduce the MMS value by 1 if the tortuosity is more than \SI{6}{\percent} higher.
These relative thresholds match the ratio of the empirical standard deviation to the mean for each metric, computed for our expert trajectories using the full dataset.
We apply comfort penalties to all trajectories except those associated with a crash or driving off-road (see \cref{tab:mms}).

To compute the similarity between planned and reference trajectories, we leverage the miss rate metric proposed by Ettinger et al.~\cite{ettinger2021large}.
We use their heuristic to calculate velocity-dependent lateral and longitudinal thresholds ($\lambda_{\text{lat}}$ and $\lambda_{\text{lon}}$).
Specifically, we calculate a threshold-based similarity with
\begin{equation}
\mathrm{sim}=
\begin{cases}
1, & \text{if } d_{\text{lat}} \le \lambda_{\text{lat}} \text{ and } d_{\text{lon}} \le \lambda_{\text{lon}}, \\
\min\!\left(\mathrm{sim}_{\text{lat}},\,\mathrm{sim}_{\text{lon}}\right), & \text{otherwise,}
\end{cases}
\label{eq:sim}
\end{equation}
where $d_{\text{lat}}$ and $d_{\text{lon}}$ are lateral and longitudinal displacements between the waypoints of the planned and reference trajectories, and $\mathrm{sim}_\text{lat}(d_\text{lat}, \lambda_{\text{lat}}) = \max\left(0, 1 - (d_\text{lat} - \lambda_{\text{lat}}) / \lambda_{\text{lat}} \right)$. 
We compute $\mathrm{sim}_\text{lon}$ analogously using the longitudinal displacement and longitudinal threshold.
The final MMS is based on 4 cases:
\begin{equation}
\label{eq:mms_plan}
\mathrm{MMS} =
\begin{cases}
0, & \text{if } \left\langle\bm{v}_\text{plan}^{(0)}, \bm{v}_\text{ref}^{(0)}\right\rangle \le 0.5\left| \bm{v}_\text{ref}^{(0)} \right|, \\
\text{MMS}_\text{ref}, & \text{else if } \text{MMS}_\text{ref} \in \{0, 1\} \text{ and } s \geq 0.4, \\
s\cdot\text{MMS}_\text{ref}, & \text{else if } s\cdot\text{MMS}_\text{ref} \geq 3.5 - \text{CP},\\ 
3.5 -\mathrm{CP}, & \text{otherwise,}
\end{cases}
\end{equation}
where $\langle\cdot,\cdot\rangle$ denotes the inner product, $\bm{v}_\text{ref}^{(0)}$ is the current reference velocity, $k$ indexes the reference trajectories, $s = \operatorname{sim}\!\big(\bm{Y}_{\text{plan}}, \bm{Y}^{(k^\star)}_{\text{ref}}\big)$ with $k^\star = \arg \max_k\operatorname{sim}\!\big(\bm{Y}_{\text{plan}}, \bm{Y}^{(k)}_{\text{ref}}\big)$, $\text{CP} \in \{0, 1, 2\}$ is the comfort penalty, and
$\text{MMS}_\text{ref}$ is the score of the most similar reference trajectory (see \cref{tab:mms}).
\begin{figure}[!t]
    \centering
    \input{figures/fig_mms_m8_row_new_new.tex}
    \caption{
    \textbf{Our pseudo-simulation via MMS covers different maneuvers and speed variations.} Specifically, we provide three reference trajectories. The expert trajectory describes the intended left turn. The wrong speed trajectory executes the same maneuver, but too slowly. The neglect instruction trajectory turns right instead of left. The matching areas at the end of each reference trajectory are based on the miss rate metric \cite{ettinger2021large}. Inside an area, \mbox{sim = 1}. Here, the predicted trajectory ends close to the wrong-speed reference's matching area (sim = 0.75). Thus, we compute the MMS by multiplying the reference MMS by the similarity.
     In this case, no comfort penalty is subtracted since jerk and tortuosity are similar to the reference trajectory.
    }
    \label{fig:mms_matching}
\end{figure}
\textbf{The first case in \cref{eq:mms_plan}} assigns planned trajectories, which are inconsistent with the past trajectory, a score of 0.
\textbf{The second case} ensures that planned trajectories, which are most similar to reference trajectories that describe crashes or driving off road (with at least moderate similarity $s \geq 0.4$), get the score of the corresponding reference trajectory.
\textbf{The third case} assigns planned trajectories, which are most similar to reference trajectories describing good or acceptable behavior (first 3 categories in \cref{tab:mms}), the score of the reference trajectory scaled by the similarity value $s$.
Additionally, we ensure that the assigned MMS value is at least as high as in the unmatched fourth case.
\textbf{The fourth case} assigns a score of 3.5 to planned trajectories, which are not matched to any reference trajectory.\footnote{We choose 3.5 as base MMS value to place this category between the \textit{neglect instruction} and \textit{driving off road} categories. We give a lower score than for the neglect instruction category, since in contrast to such reference trajectories, unmatched trajectories can neglect traffic rules. We give a higher score than for the driving off road and crash categories, since (1) unmatched trajectories are at least consistent with the past trajectory (not case 1 in \cref{eq:mms_plan}) and (2) unmatched trajectories are not similar to the explicit cases of driving off road or crashing (represented by the labeled reference trajectories).}
As in the previous case, we also subtract comfort penalties (CP) based on jerk and tortuosity.

\subsubsection{Fallback logic for non-parseable or copied trajectories.} Inspired by the baseline in NAVSIM \cite{dauner2024navsim}, we default to a constant velocity trajectory with constant heading if trajectories are not parseable in model outputs or if models just copy-paste the past trajectory into the output. 

\subsection{Zero-shot inference setting}
\label{subsec:zero_shot_inference}
\subsubsection{Native reasoning versus prompted reasoning.} We evaluate all models zero-shot, using publicly available checkpoints or APIs.  
For the driving-specific reasoning models (AutoVLA, OneVL, Alpamayo 1.5, Alpamayo 2), we use their native reasoning.
These models are trained to generate reasoning traces as part of their inference procedure, so we run them in their default configuration without prompting them specifically to reason about driving actions. 
For the embodied and general-purpose reasoning models, which have no built-in trajectory decoding, we elicit reasoning via chain-of-thought prompting, specifically the zero-shot instruction-based variant \cite{kojima2022large} with a specific reasoning structure requested before generating a future trajectory. 
Concretely, the prompt asks the model to first state its situational awareness and the reasons behind its intended steering and acceleration commands, and only then to output its answer in our structured format (see \cref{fig:reasoned_actions_vs_executed_actions}). 
The full prompt is included in the extended data section (see \cref{subsec:zero_shot_cot_prompt}).
Our setup compares the native reasoning of domain-specific models against prompt-elicited reasoning on top of general-purpose reasoning capabilities, while keeping the downstream evaluation of coherence and motion planning identical across all models.
\subsubsection{Input data.} The driving-specific models receive all ring camera images, while the embodied and general-purpose models receive the front-view image of the current time step. 
All models additionally receive the past trajectory and the high-level instruction as context. 

\subsection{Making reasoning the action interface: coherence by construction}
\label{sec:kinematic_model}


We propose a command-conditioned kinematic bicycle model that maps natural language acceleration and steering commands to continuous control inputs for generating trajectories.  
We parameterize the kinematic bicycle model using our research vehicle's geometry, with a wheelbase of 3.21~\si{\metre} and a maximum steering angle of $33.83^\circ$.
To convert discrete language commands into continuous magnitudes, we tune the control values associated with each acceleration and steering command. 
Since the same command can require different control magnitudes at different speeds, we employ separate parameterizations below and above 60~km/h.
We tune these parameters against expert trajectories using a two-stage constrained optimization procedure on our training split, and validate the resulting parameterization on the validation split. 
First, we optimize the command-to-control mapping to ensure that trajectories generated from a given set of commands are classified back into the same acceleration and steering categories, thereby enforcing reasoning-action coherence. 
We then refine the feasible parameters by minimizing trajectory displacement error while preserving this command consistency.
Further implementation details are in \cref{subsec:tuning_the_kinematic_bicycle_model}.

\subsection{Dataset specifics}

\label{sec:dataset}
We collected our data over the course of two years, beginning in late 2023. 
Our recordings include urban and suburban environments, as well as highways
(the main locations are listed in \cref{tab:e2e_benchmarks}).
We adjusted our routes to include many construction zones and intersections. 
In particular, we filtered for rare events such as adverse weather conditions, road closures, and accidents.
Consequently, our dataset encompasses scenarios that diverge from nominal data distributions (i.e., long-tail scenarios).
Overall, our dataset contains one thousand \SI{9}{\second}-long scenarios that are divided into three splits: train (\num{500}), test (\num{400}), and validation (\num{100}).

\subsubsection{Scenario types.}
\begin{figure}[htbp]
    \centering
    \includegraphics[width=\linewidth]{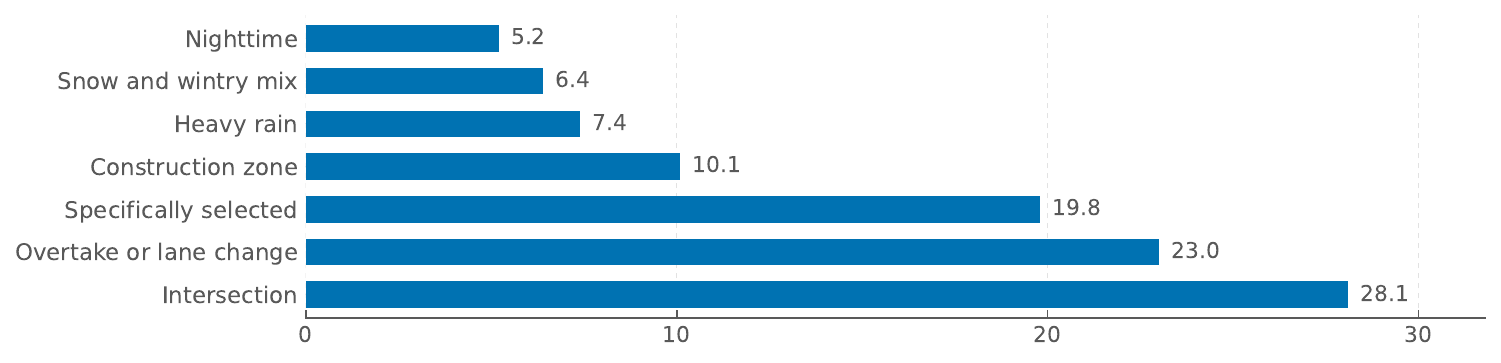}
    \caption{\textbf{Distribution of scenario types.} Numbers are percentages.}
    \label{fig:scenario_distribution}
\end{figure}

\cref{fig:scenario_distribution} shows the distribution of scenario types, which is approximately equal across our train, val, and test split.
In addition to specifically selected challenging scenarios (road closures, accidents, or combinations of the other scenario types), adverse weather, and construction zones, we use the Pareto principle to determine further long-tail data.
Specifically, we use the well-established nuScenes dataset \cite{caesar2020nuscenes} as reference and rank-frequency plots with an 80\% cumulative frequency threshold.
In nuScenes, approx. 88\% of the scenarios are recorded during the day, thus nighttime scenarios are long-tail data.
For maneuver types,  driving straight and regular turns account for approx. 90\% of nuScenes.
Furthermore, we include overtaking and lane changing at high speeds ($> 100\si{\km/\hour}$), whereas the maximum speed in nuScenes is approx. $60\si{\km/\hour}$.
Therefore, our overtaking and lane changing scenarios are long-tail scenarios. 
As an exception, we also include nominal driving at intersections to better evaluate instruction following since there are more viable trajectories than in most long-tail scenarios.

\subsubsection{Video data.}
Our dataset contains multi-view video data with a \ang{360} horizontal field of view (FoV) and six viewing angles (see top row in \cref{fig:multiview_video_data}).
For the corresponding frames, we provide two image formats: raw (image size of $3200 \times 2200 \times 3$) and pinhole (image size of  $3488\times 2272 \times 3$), based on a non-single viewpoint and a pinhole camera model.
We optimize the pinhole parameters to create images that can be processed as $16 \times 16\,\si{px}$ patches (see ViTs \cite{dosovitskiy2020image}).

\subsubsection{LiDAR-calibrated trajectories.}
We extract 3D poses from LiDAR data using the LiDAR odometry pipeline KISS-ICP \cite{vizzo2023ral}.
Afterwards, we project the poses into bird's-eye-view and smooth the corresponding trajectories using an extended Kalman Filter.

\begin{figure}[h]
    \centering
    \input{figures/multi-view/fig2}
    \caption{
    \textbf{
    We provide multi-view video data and past  motion trajectories, and prompt reasoning models to reason before planning a future trajectory. 
    }
    The reasoning covers situational awareness and driving actions (i.e., steering and acceleration).
    Our reasoning evaluation (see \cref{subsec:reasoning_and_action_frequently_disagree}) is split into two segments: 4\,s to 7\,s (first 3\,s of the future trajectory) and 7\,s to 9\,s (last 2\,s of the future trajectory).
    We adapt this structure from well-established motion forecasting benchmarks \cite{ettinger2021large, wilson2023argoverse}.
    }
    \label{fig:multiview_video_data}
\end{figure}

\subsubsection{High-level instructions.}

We provide high-level driving instructions that describe the intended maneuver in each scenario.
All instructions were manually annotated by domain experts.
The most common command type is \emph{drive straight on} (\SI{\fpeval{373/826*100}}{\percent}), followed by turn maneuvers \emph{turn right} (\SI{\fpeval{120/826*100}}{\percent}), \emph{turn left} (\SI{\fpeval{51/826*100}}{\percent}) and use lane instructions such as \emph{use right lane} (\SI{\fpeval{64/826*100}}{\percent}) or \emph{use left lane} (\SI{\fpeval{54/826*100}}{\percent}).
A distinctive feature of the dataset is the detailed formulation of overtake commands (\SI{\fpeval{112/826*100}}{\percent}), which often specify both the object type and its relative position, for example \emph{overtake truck driving on the right} or \emph{overtake car in front}.
In general, instructions can be followed throughout the scenario, but in many \emph{specifically selected} cases this is intentionally not possible.
In these scenarios, the instructed maneuver cannot be executed due to external factors such as oncoming traffic, obstacles, or the ego vehicle itself \mbox{being overtaken}.
Compared to purely route-based directives, such as \emph{left}, \emph{right}, or \emph{straight}, used in benchmarks like Bench2Drive \cite{jia2024bench2drive} or Waymo Open E2E \cite{waymo2025e2e}, these fine-grained textual instructions enable a more precise evaluation of instruction following and context-aware decision making.

\subsubsection{Our dataset and the corresponding evaluation fill a benchmark gap in end-to-end autonomous driving.}
End-to-end driving methods~\cite{hu2023planning, jiang2023vad, sun2025sparsedrive, hwang2025emma, rowe2025poutine, sima2024drivelm} are fully differentiable models that take raw sensor data (e.g., video, LiDAR, radar, or GNSS data) as input and output planned ego trajectories.
Despite its limitations (see \cite{li2024ego}), benchmarking of such methods on nuScenes \cite{caesar2020nuscenes} is still common (e.g.,~\cite{hwang2025emma, sun2025sparsedrive, zhang2025future}).
The corresponding evaluation protocol of Hu et al.~\cite{hu2023planning} computes the $L_2$ error with respect to an expert trajectory and collision rates with other road users.
Thus, the evaluation is non-reactive and considers only one maneuver as ground truth.
To consider multiple possible maneuvers, NAVSIM~\cite{dauner2024navsim} builds upon nuPlan \cite{caesar2021nuplan} and introduces non-reactive simulation metrics.
This includes metrics like progress and time to collision, but simulated ego trajectories and environments do not influence each other.
Bench2Drive~\cite{jia2024bench2drive} is an end-to-end driving benchmark that builds upon the CARLA simulator~\cite{dosovitskiy2017carla}.
Its metrics like success rate and DrivingScore are based on reactive simulation. 
However, simulated sensor data exhibits a large domain gap to real data.
The Waymo Open E2E benchmark~\cite{waymo2025e2e} evaluates end-to-end driving methods on rare long-tail scenarios, including construction zones, foreign object debris, or special vehicles.
At the time of this writing, they do not provide video data, but just the camera images for the current time step.
Furthermore, the benchmark data does not include reasoning traces and semantic reasoning-action coherence of model outputs is not evaluated.
We list further details on benchmarks and datasets for end-to-end driving in \cref{tab:e2e_benchmarks}.

\begin{table*}[h]
    \centering
    \normalsize
    \resizebox{\textwidth}{!}{
    \begin{tabular}{lcccScccl}
        \toprule
        \multirow{2}{*}{Dataset} & Reasoning-action & Long-tail & Expert & Planning & Multi-maneuver & Real video & High-level & \multirow{2}{*}{Main locations} \\
                & coherence & data      & reasoning & {horizon [\si{\second}]} & evaluation & data & instructions & \\
        \midrule
        nuScenes \cite{caesar2020nuscenes} & \ding{55} & \ding{55} & \ding{55} & 3 & \ding{55} & \checkmark & $\halfcirc$ & Boston, Singapore \\
        NAVSIM \cite{dauner2024navsim} & \ding{55} & \ding{55} & \ding{55} & 4 & \checkmark & \checkmark & $\halfcirc$ & Boston, Singapore \\
        Bench2Drive \cite{jia2024bench2drive} & \ding{55} & $\halfcirc$ & \ding{55} & {varying} & \checkmark & \ding{55} & \checkmark & CARLA cities (simulation) \\
        Waymo Open E2E \cite{waymo2025e2e} & \ding{55} & \checkmark & \ding{55} & 5 & \checkmark & \ding{55} & $\halfcirc$ & 12 U.S. cities\\
        DriveLM-Data \cite{sima2024drivelm} & \ding{55} & \ding{55} & $\halfcirc$ & 3 & \ding{55} & $\halfcirc$ & $\halfcirc$ & Boston, Singapore, CARLA \\
        CoVLA-Dataset \cite{arai2025covla} & \ding{55} & $\halfcirc$ & \ding{55} & 3 & \ding{55} & \checkmark & \checkmark & Tokyo \\ \midrule
        Our dataset & \checkmark & \checkmark & \checkmark & 5 & \checkmark & \checkmark & \checkmark & Karlsruhe, Heidelberg, \\
        & & & & & & & &  Mannheim, Black Forest\\
        \bottomrule
    \end{tabular}}
    \caption{\textbf{Our dataset and evaluation fill a benchmark gap and target reasoning-capable end-to-end driving methods.} A half-filled circle indicates that a feature is partially available. For example, regarding long-tail scenarios, related work selects interesting scenarios based on variations in trajectories instead of scenario classes such as navigating a construction zone. As high-level instructions, related work provides only \{right, left, straight\}.}
    \label{tab:e2e_benchmarks}
\end{table*}

\clearpage  

\section*{Acknowledgements}
The research leading to these results is partially funded by the German Federal Ministry for Economic Affairs and Energy within the project ``NXT GEN AI METHODS''.
The authors gratefully acknowledge the computing time provided on the high-performance computer HoreKa by the National High-Performance Computing Center at KIT (NHR@KIT). This center is jointly supported by the Federal Ministry of Education and Research and the Ministry of Science, Research and the Arts of Baden-Württemberg, as part of the National High-Performance Computing (NHR) joint funding program. HoreKa is partly funded by the German Research Foundation (DFG).
J. Villa acknowledges funding from the predoctoral contract PRE2022-104690, funded by the Spanish State Research Agency (AEI, MCIN/AEI/10.13039/501100011033) and co-funded by the European Social Fund Plus (FSE+).
Last but not least, we thank Florian Wirth for supporting the work in its early stages.

%
%
\bibliographystyle{splncs04}
\bibliography{main}
\clearpage

\section{Extended data} 

\subsection{Qualitative results}

\begin{figure*}[!h]
    \vspace{-0.5cm}
    \centering
    \begin{subfigure}[t]{0.49\textwidth}
        \centering
        \includegraphics[width=\linewidth]{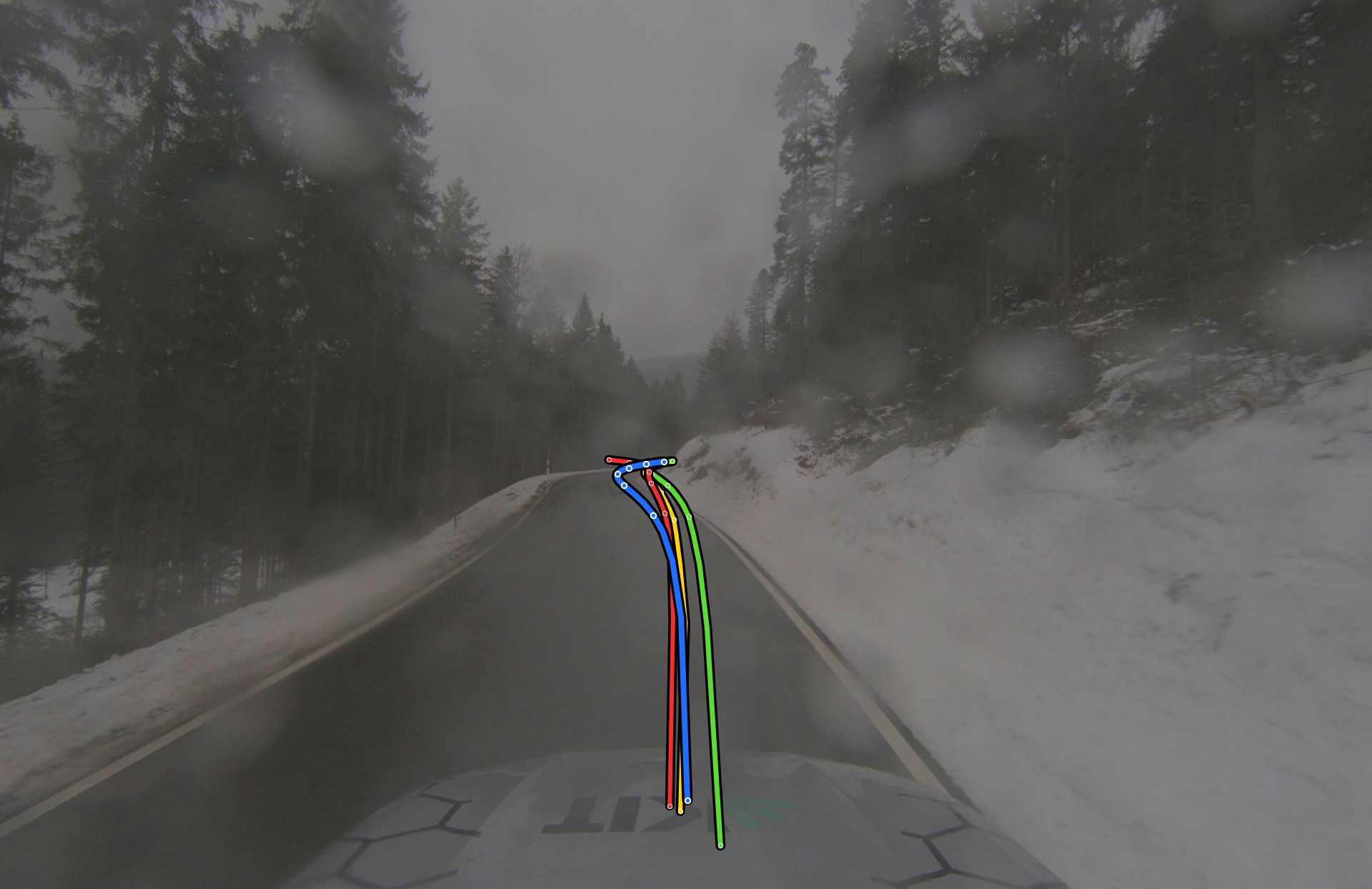}
        \caption{\textbf{Scenario type:} Snow and wintry mix; \textbf{Instruction:} Drive straight on; \textbf{MMS:} 10.0; \textbf{Model:} Alpamayo 2 \cite{nvidia2026alpamayo2}}
    \end{subfigure}
    \begin{subfigure}[t]{0.49\textwidth}
        \centering
        \includegraphics[width=\linewidth]{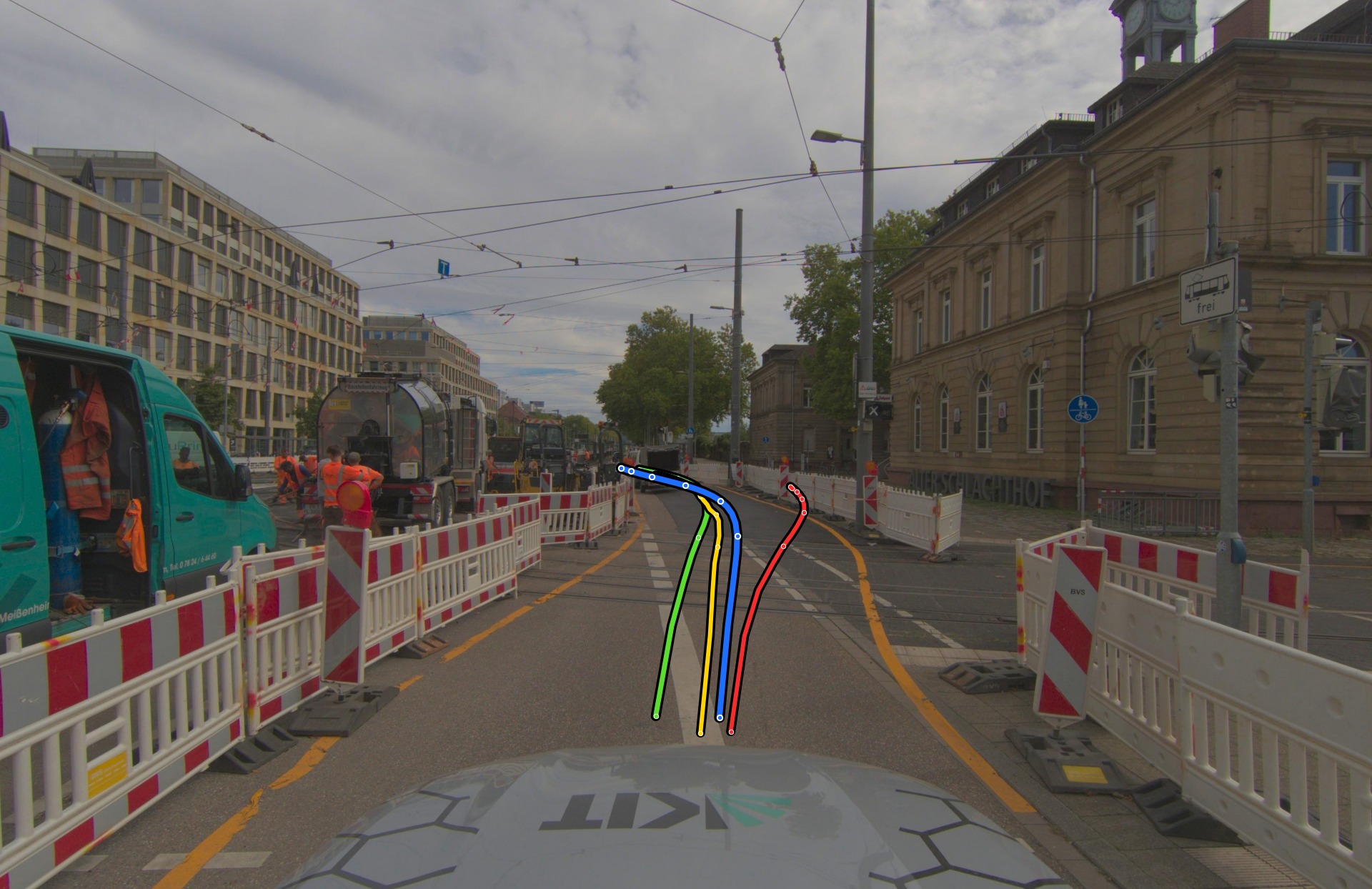}
        \caption{\textbf{Scenario type:} Construction zone; \textbf{Instruction:} Drive straight on; \textbf{MMS:} 10.0; \textbf{Model:} Alpamayo 2 \cite{nvidia2026alpamayo2}}
    \end{subfigure}
    \vspace{3pt}
    
    \begin{subfigure}[t]{0.49\textwidth}
        \centering
        \includegraphics[width=\linewidth]{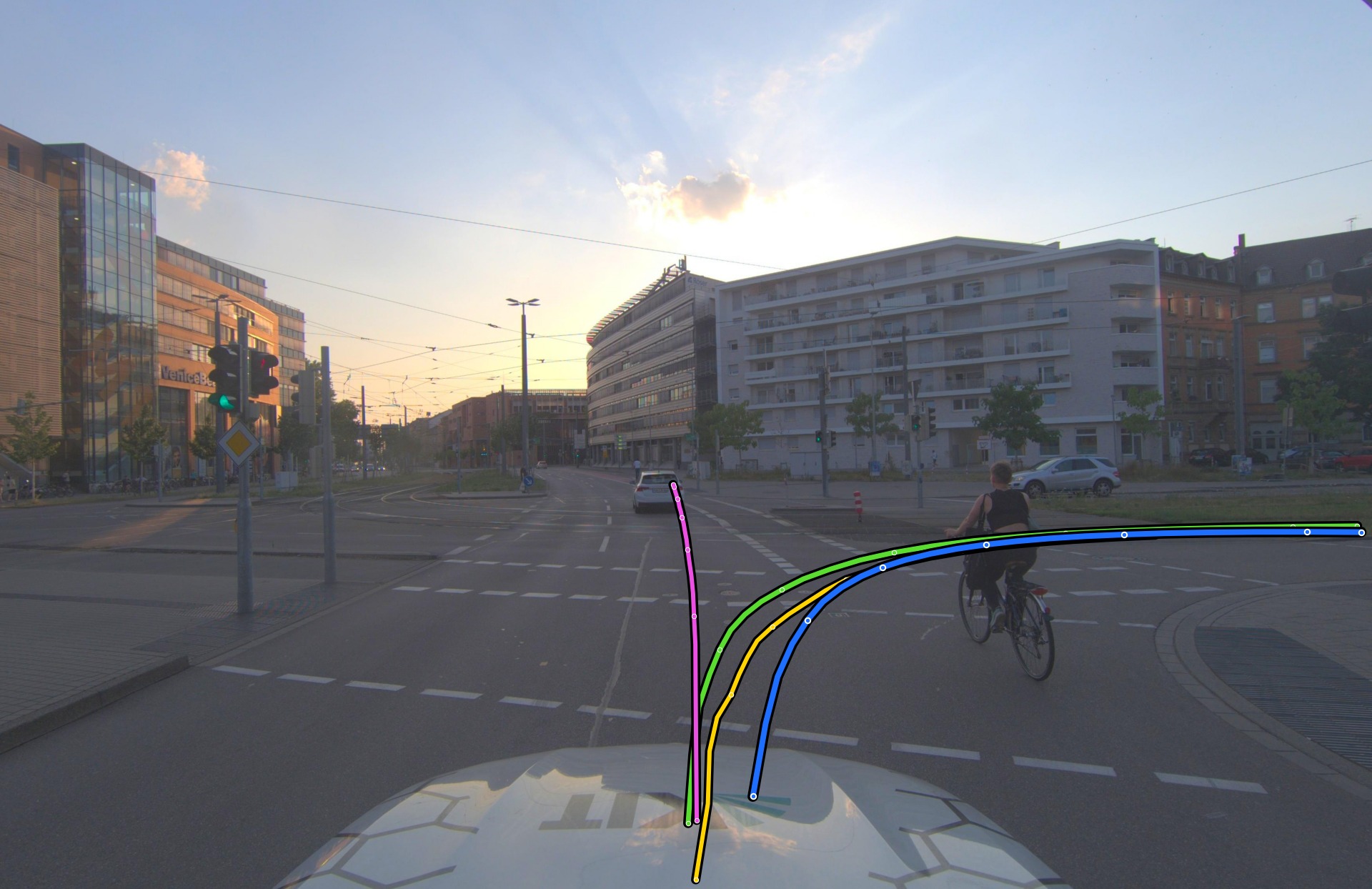}
        \caption{\textbf{Scenario type:} Intersection; \textbf{Instruction:} Turn right; \mbox{\textbf{MMS:} 10.0}; \textbf{Model:} Gemini Robotics ER 1.6 \cite{team2025gemini}}
    \end{subfigure}
    \begin{subfigure}[t]{0.49\textwidth}
        \centering
        \includegraphics[width=\linewidth]{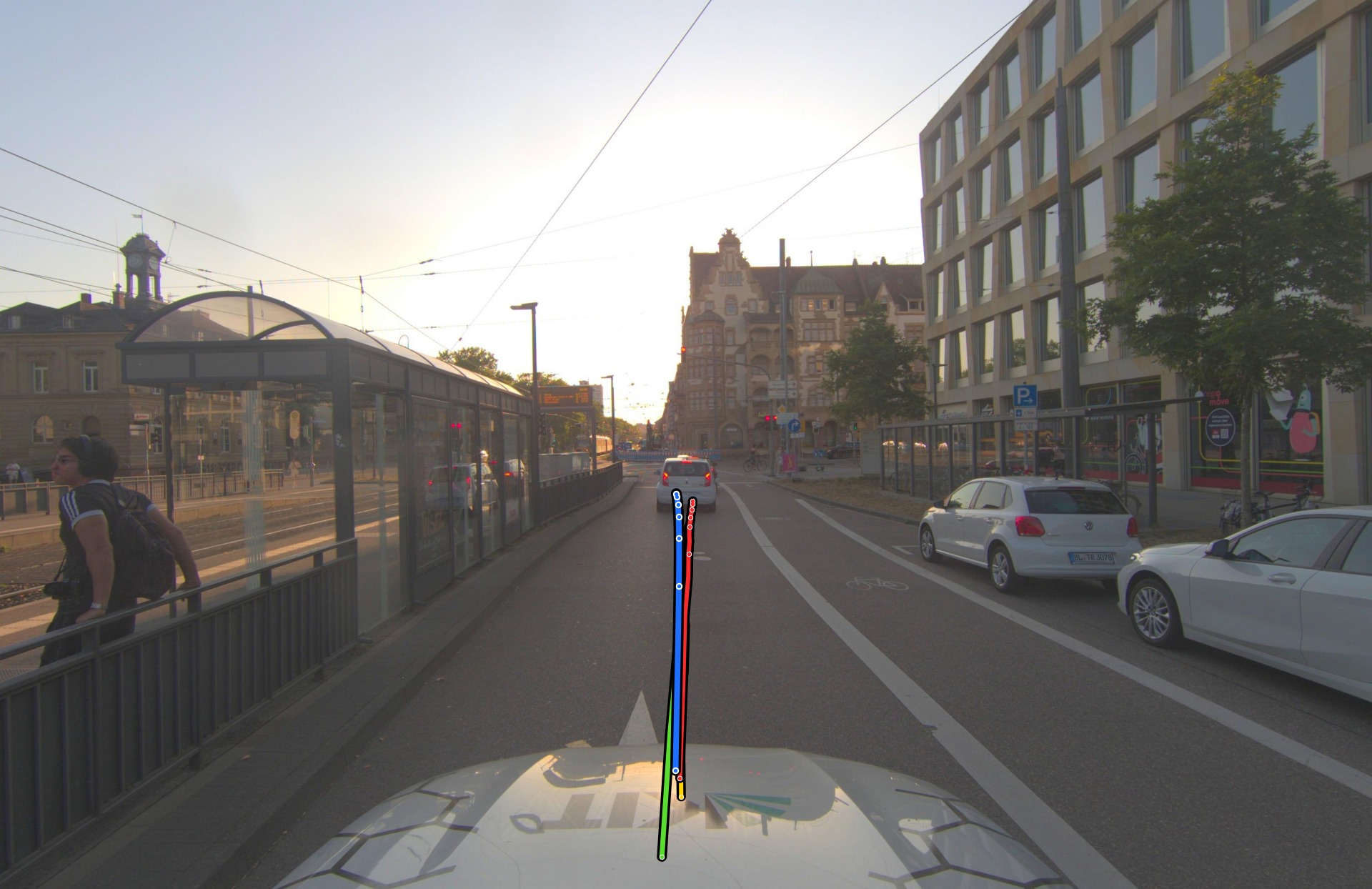}
        \caption{\textbf{Scenario type:} Intersection; \textbf{Instruction:} Drive straight on; \textbf{MMS:} 0.0; \textbf{Model:} Gemini Robotics ER 1.6 \cite{team2025gemini}}
    \end{subfigure}
    \vspace{3pt}
    
    \begin{subfigure}[t]{0.49\textwidth}
        \centering
        \includegraphics[width=\linewidth]{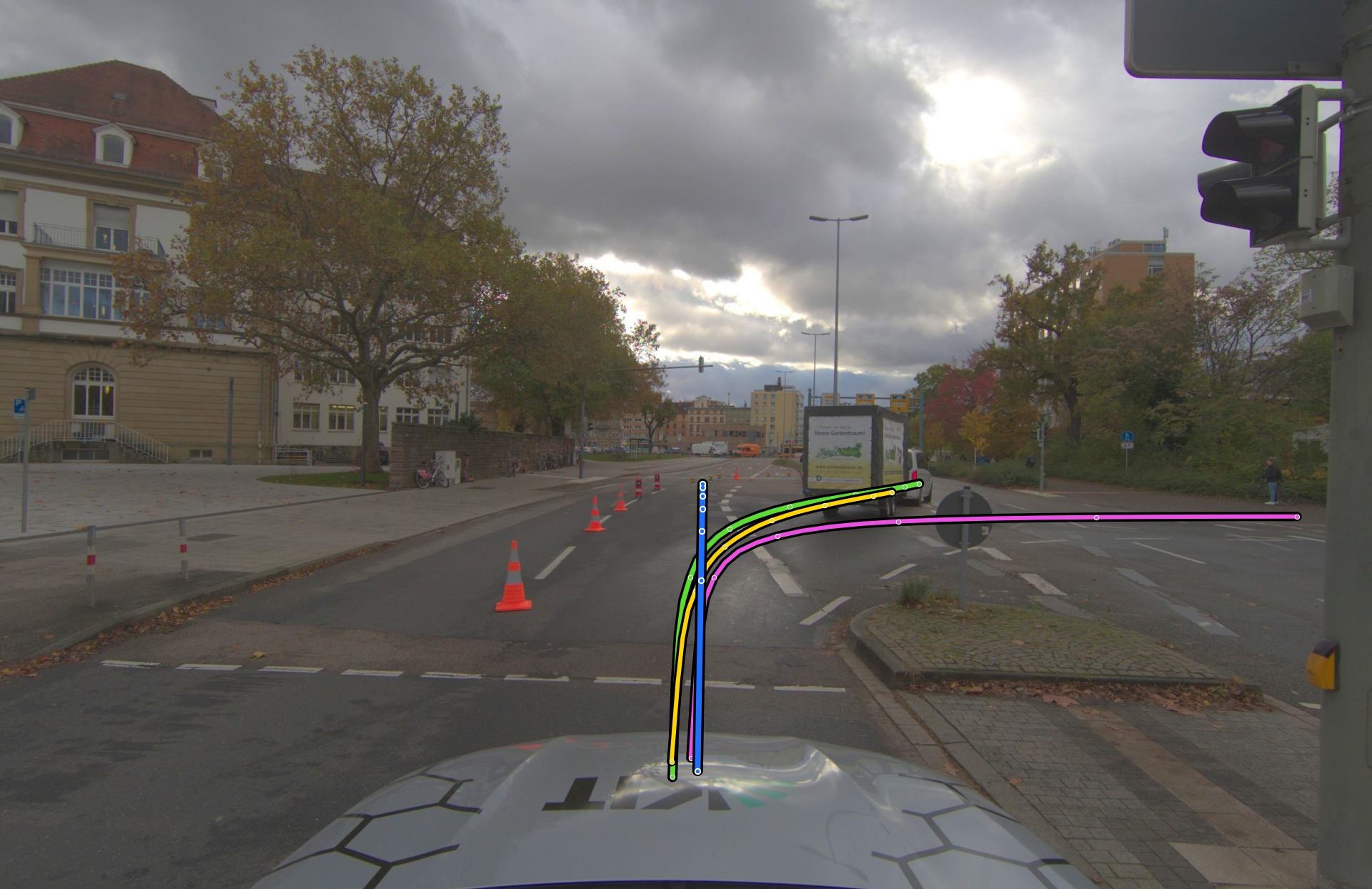}
        \caption{\textbf{Scenario type:} Specifically selected; \textbf{Instruction:} Drive straight on; \textbf{MMS:} 3.5; \textbf{Model:} GPT-5.6 Luna \cite{singh2025openai}}
    \end{subfigure}
    \begin{subfigure}[t]{0.49\textwidth}
        \centering
        \includegraphics[width=\linewidth]{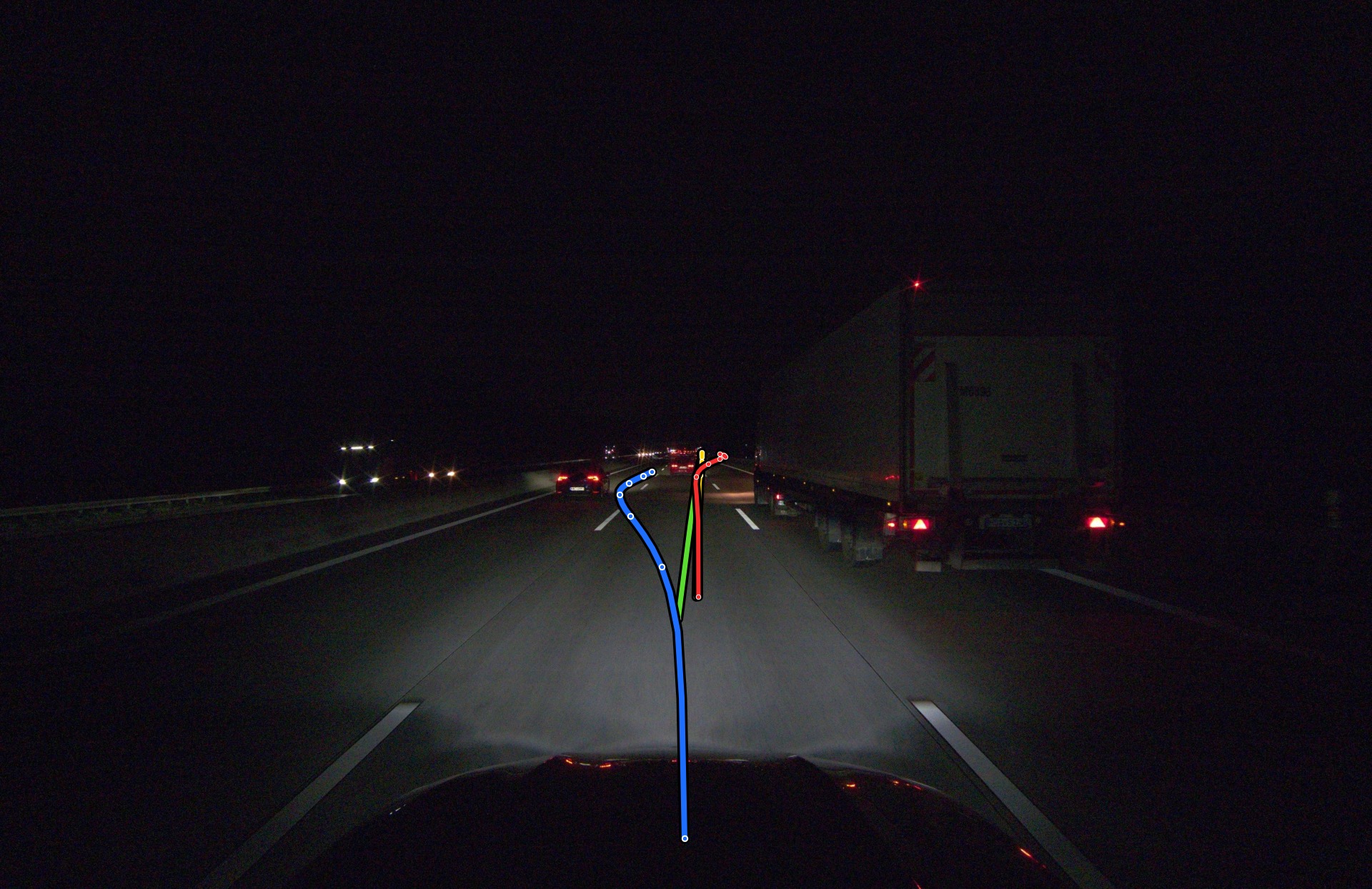}
        \caption{\textbf{Scenario type:} Overtake; \textbf{Instruction:} Overtake truck driving on the right; \textbf{MMS:} 3.5; \textbf{Model:} GPT-5.6 Luna \cite{singh2025openai}}
    \end{subfigure}
    \caption{\textbf{Qualitative results of driving-specific (Alpamayo 2), embodied (Gemini Robotics ER 1.6) and general-purpose (GPT-5.6 Luna) reasoning models.}
    \textbf{(a) to (c):} Show examples where the predicted trajectory (blue) is perfectly matched to the expert trajectory (green).
    \textbf{(d)}: Shows an example where the predicted trajectory is matched to a crash trajectory (red, representing a rear-end crash).
    \textbf{(e) and (f):} Examples where the predicted trajectory is not matched to any reference trajectory. Additionally, neglect instruction (i.e., different maneuver) and wrong speed trajectories are shown in pink and yellow.
    }
    \label{fig:qualitative_results}
\end{figure*}

\subsection{Scenario examples}
\begin{figure*}[!h]
    \centering
    \begin{subfigure}[t]{0.49\textwidth}
        \centering
        \includegraphics[width=\linewidth]{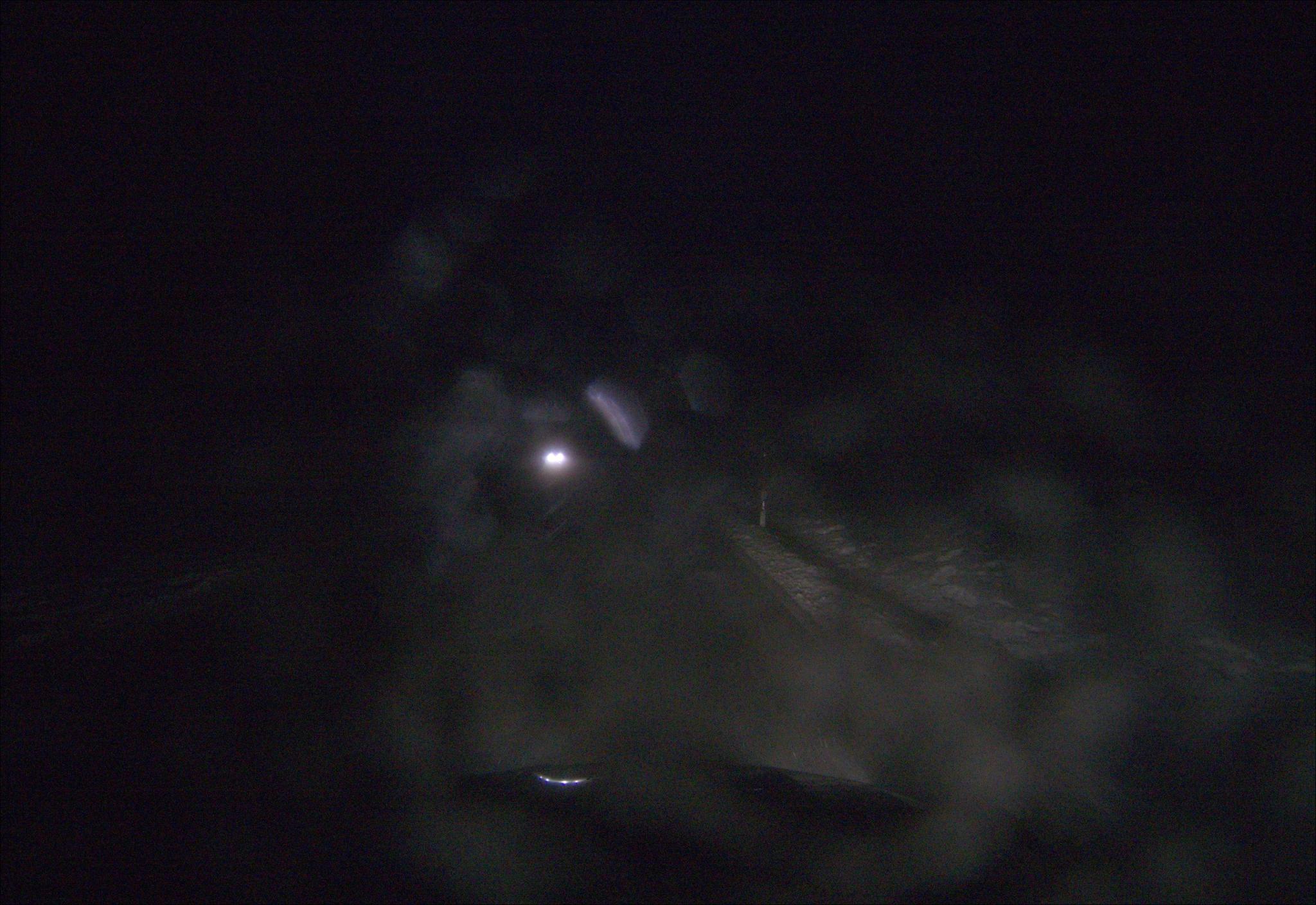}
        \caption{\textbf{Scenario type:} Specifically selected}
    \end{subfigure}
    \begin{subfigure}[t]{0.49\textwidth}
        \centering
        \includegraphics[width=\linewidth]{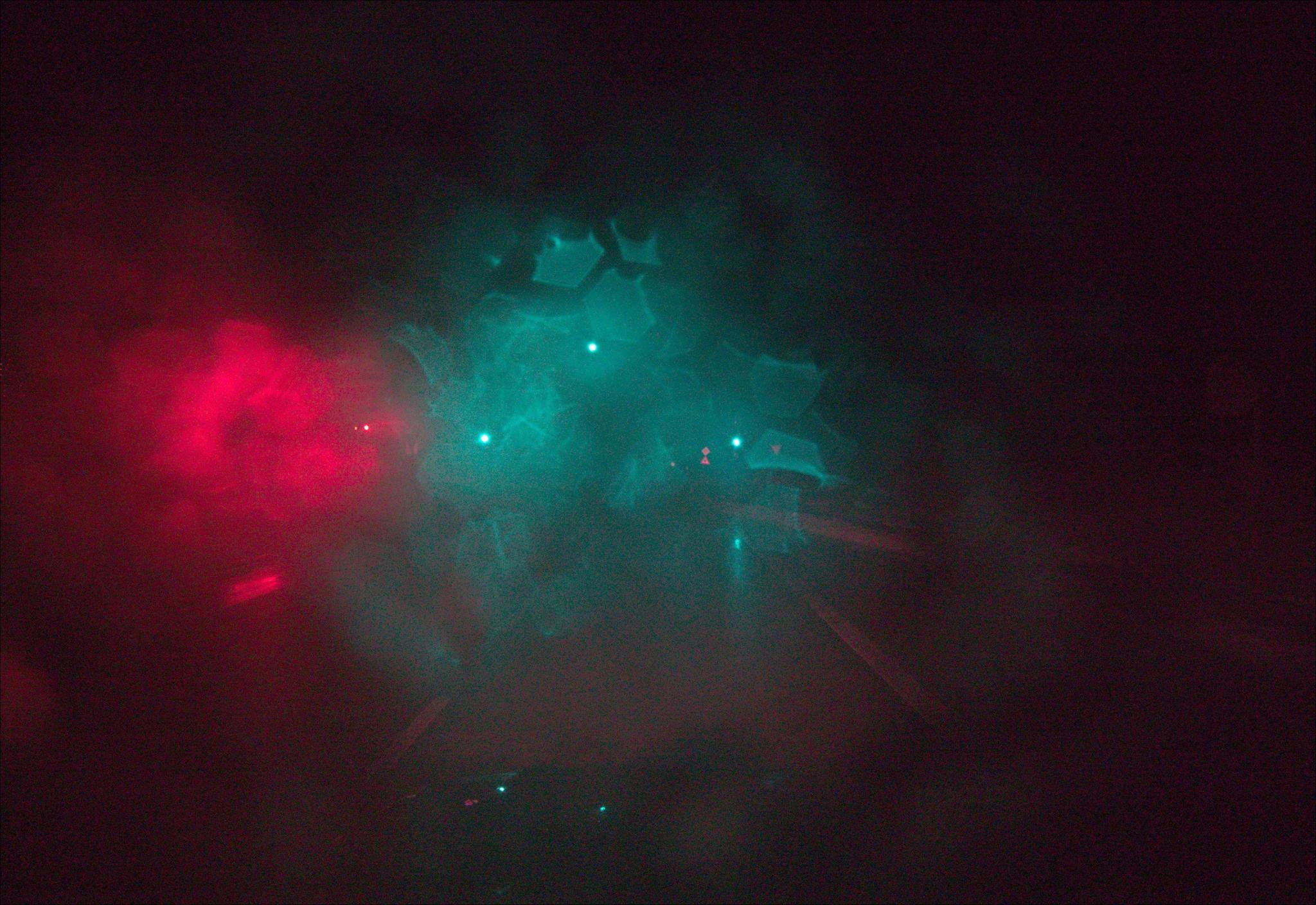}
        \caption{\textbf{Scenario type:} Specifically selected}
    \end{subfigure}
    \vspace{3pt}

    \begin{subfigure}[t]{0.49\textwidth}
        \centering
        \includegraphics[width=\linewidth]{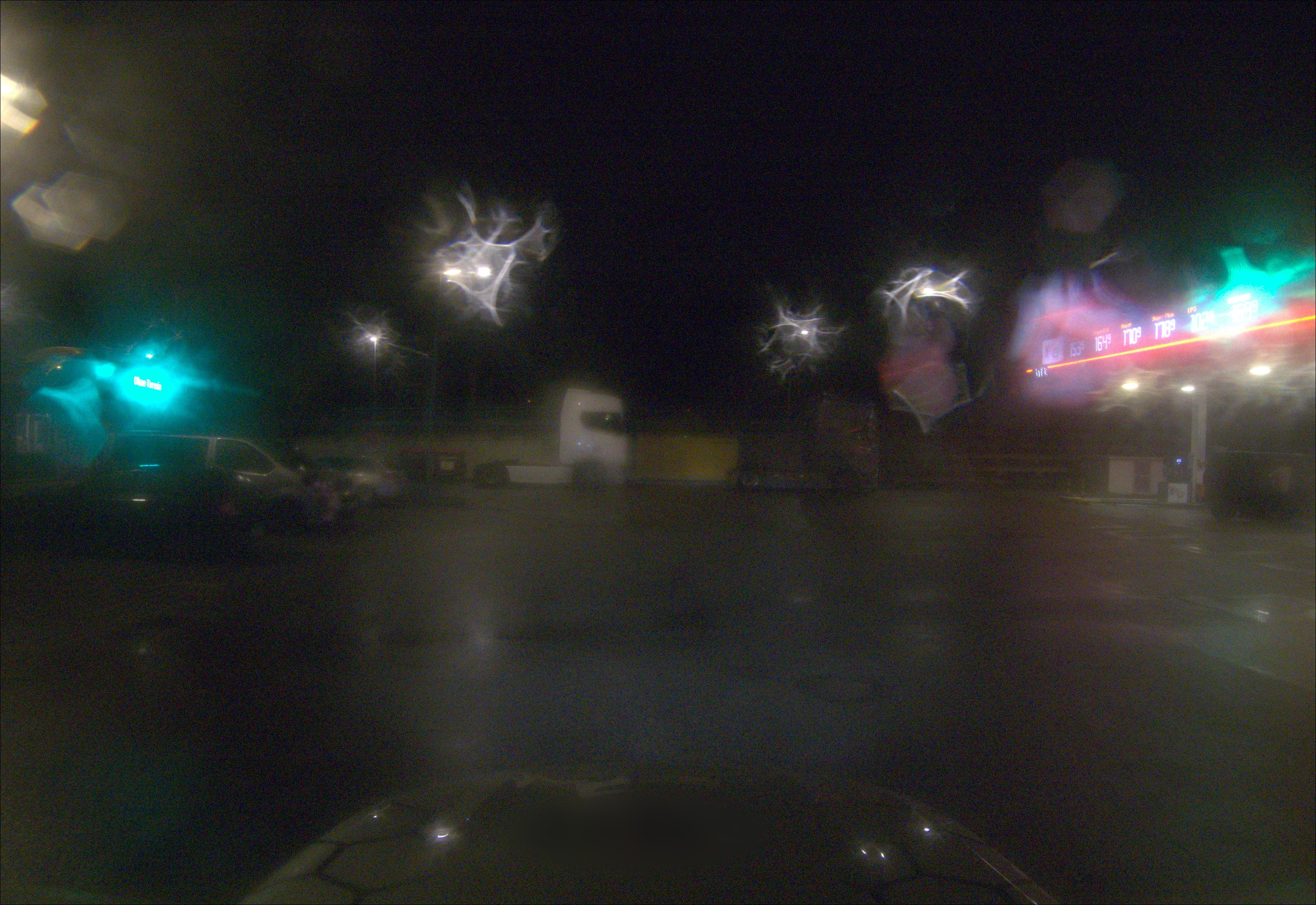}
        \caption{\textbf{Scenario type:} Specifically selected}
    \end{subfigure}
    \begin{subfigure}[t]{0.49\textwidth}
        \centering
        \includegraphics[width=\linewidth]{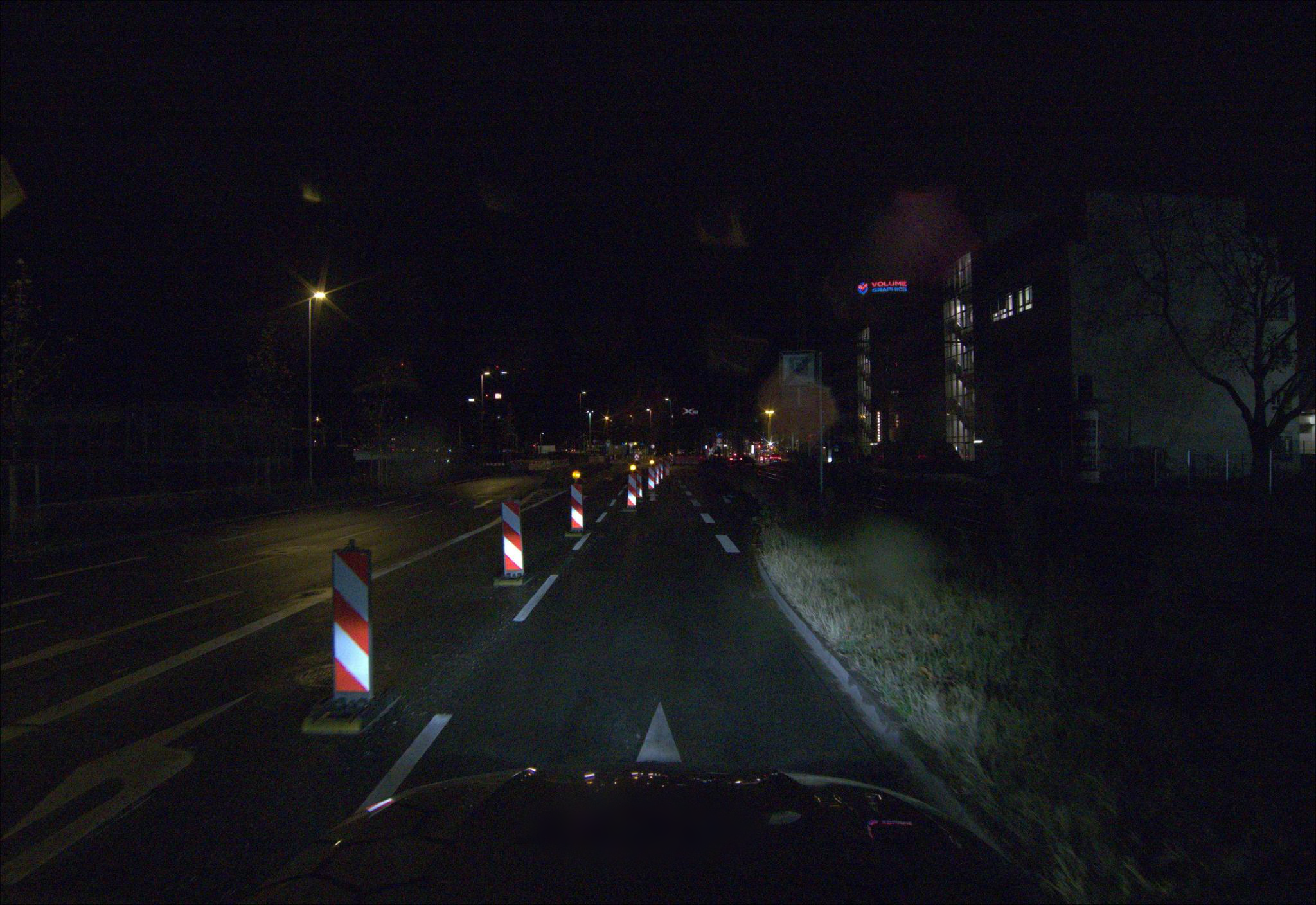}
        \caption{\textbf{Scenario type:} Specifically selected}
    \end{subfigure}
    \vspace{3pt}

    \begin{subfigure}[t]{0.49\textwidth}
        \centering
        \includegraphics[width=\linewidth]{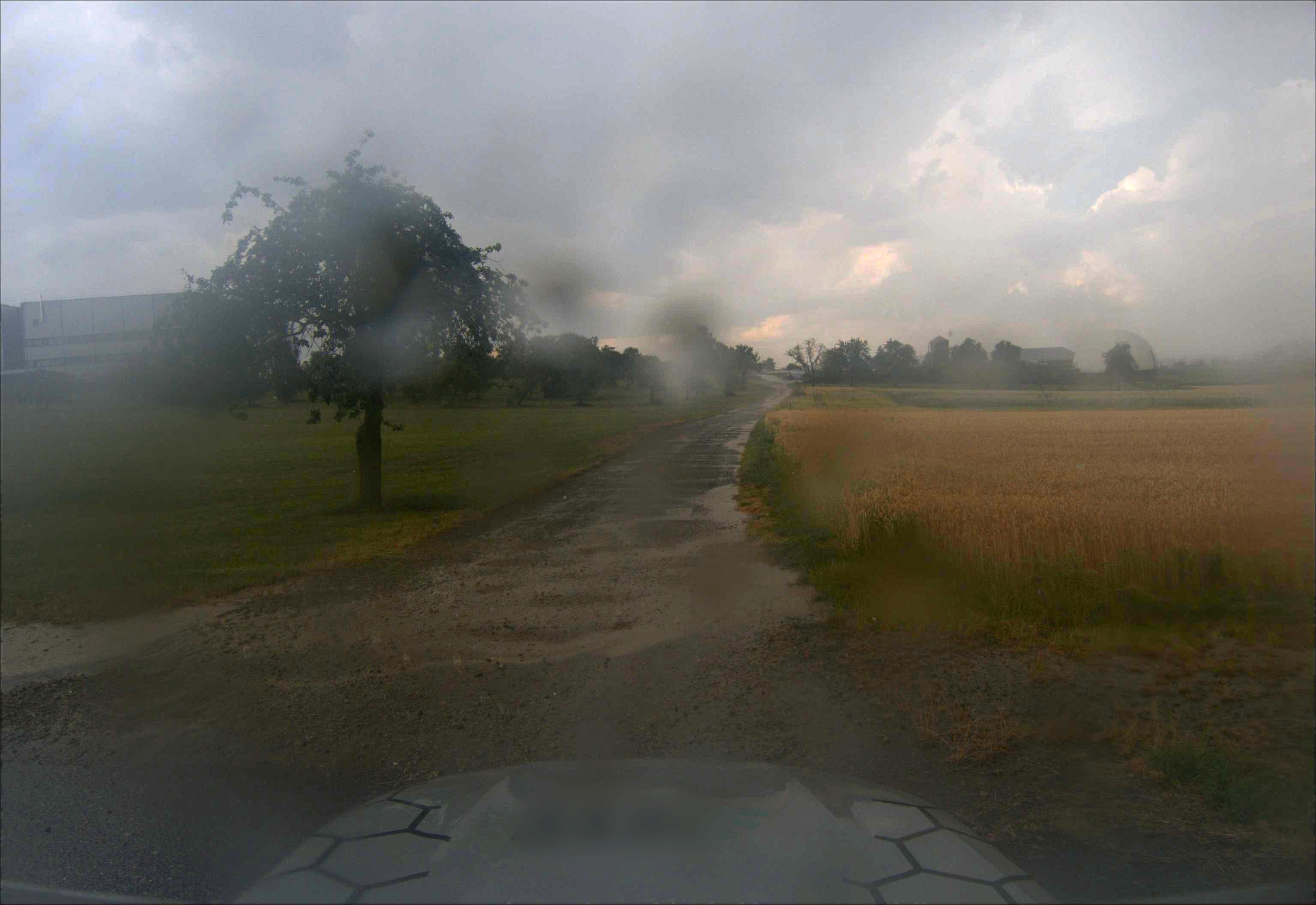}
        \caption{\textbf{Scenario type:} Heavy rain}
    \end{subfigure}
    \begin{subfigure}[t]{0.49\textwidth}
        \centering
        \includegraphics[width=\linewidth]{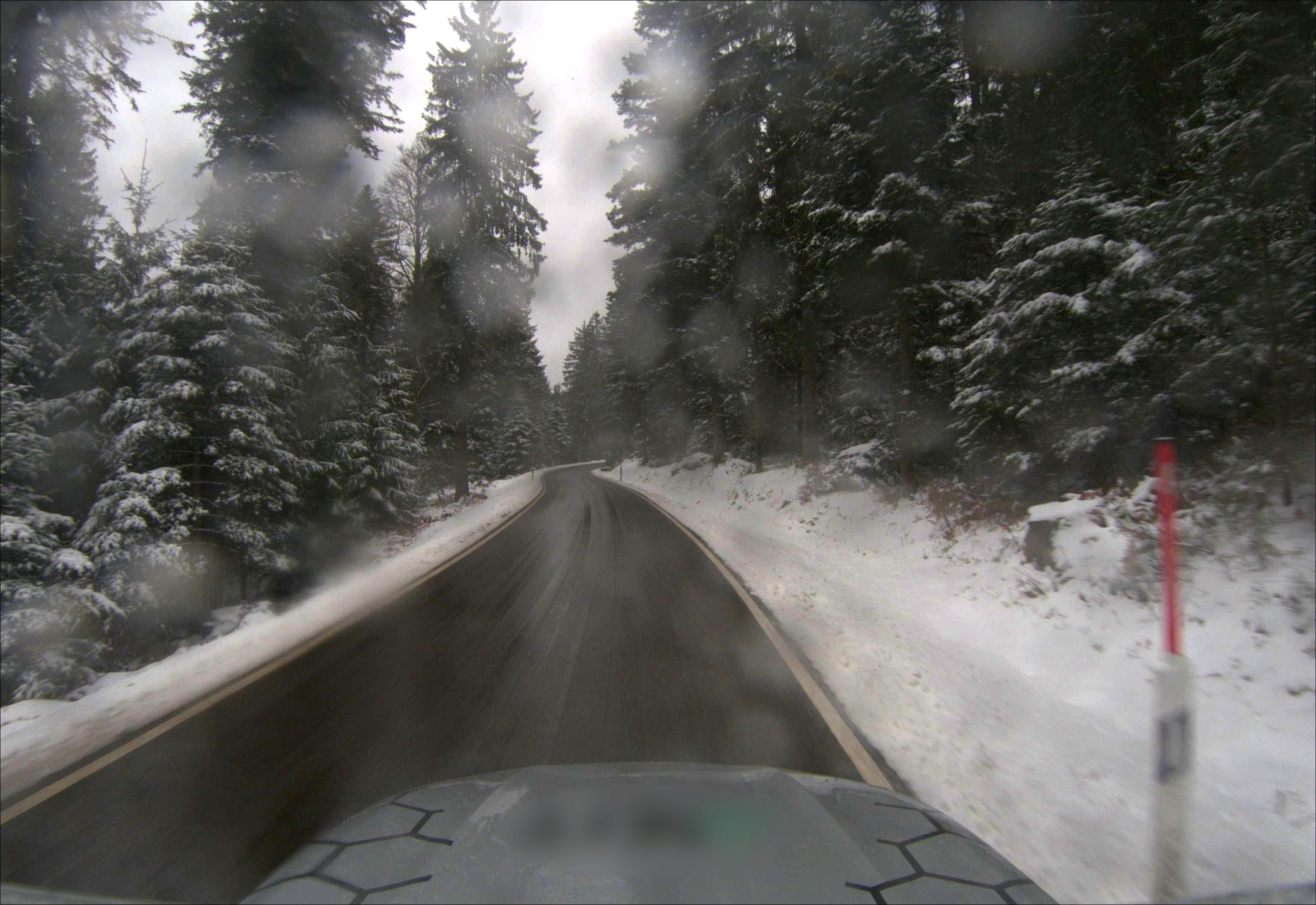}
        \caption{\textbf{Scenario type:} Snow and wintry mix}
    \end{subfigure}
    \vspace{3pt}
    \caption{\textbf{Front-view images of specifically selected, heavy rain, and snow scenarios.}
    In addition to rare events like protesting climate activists (shown in the main paper), crashes, or road closures we also specifically select combinations of other long-tail classes. For example:
    \textbf{(a)} Specifically selected because of wintry mix and during the night. 
    \textbf{(b) and (c)} Specifically selected because of heavy rain and during the night. 
    \textbf{(d)} Specifically selected because of a construction zone and during the night. 
    }
    \label{fig:scenario_examples}
\end{figure*}

\subsection{Zero-shot CoT prompt for general-purpose and embodied reasoning models}
\label{subsec:zero_shot_cot_prompt}
\begin{prompt}[width=1.0\linewidth]{Zero-shot CoT prompting example}{We provide front-view images, the past trajectory, and driving instructions as context and describe the allowed driving actions and formatting requirements.}{prompt:fewshot_cot_kinematic_prompt}
\raggedright
\footnotesize
\$IMAGE\_PATH\$
\par
\texttt{<past\_trajectory>}
(-123.42, -4.99), (-117.22, -4.75), (-111.02, -4.52), (-104.81, -4.28), (-98.65, -4.03), (-92.46, -3.78), (-86.29, -3.51), (-80.14, -3.21), (-73.95, -2.94), (-67.79, -2.64), (-61.63, -2.32), (-55.47, -2.05), (-49.31, -1.76), (-43.14, -1.49), (-36.97, -1.22), (-30.82, -0.98), (-24.68, -0.76), (-18.5, -0.56), (-12.33, -0.37), (-6.16, -0.17), (0.0, 0.0)
\texttt{</past\_trajectory>}
\par
\texttt{<driving\_instruction>}use right lane\texttt{</driving\_instruction>}
\par
\texttt{<task>}
Imagine you are driving the car in the image. Based on the front-view image, the past trajectory recorded at 5Hz, and the driving instruction, generate acceleration and steering commands for a 5s-long future trajectory.
\vspace{2mm}\par
The only allowed acceleration commands are: \\
- accelerate slightly\\
- accelerate strongly\\
- maintain the current speed\\
- decelerate slightly\\
- decelerate strongly\\
\vspace{2mm}\par
The only allowed steering commands are:\\
- steer slightly to the left\\
- steer to the left\\
- steer straight\\
- steer slightly to the right\\
- steer to the right\\

\vspace{2mm}\par
Your XML output must follow **exactly** this structure and tag order:
\texttt{<situational\_awareness>}...\texttt{</situational\_awareness>}\\
\texttt{<acceleration\_first\_3s>}...\texttt{</acceleration\_first\_3s>}\\
\texttt{<reason\_acceleration\_first\_3s>}...\texttt{</reason\_acceleration\_first\_3s>}\\
\texttt{<steering\_first\_3s>}...\texttt{</steering\_first\_3s>}\\
\texttt{<reason\_steering\_first\_3s>}...\texttt{</reason\_steering\_first\_3s>}\\
\texttt{<acceleration\_last\_2s>}...\texttt{</acceleration\_last\_2s>}\\
\texttt{<reason\_acceleration\_last\_2s>}...\texttt{</reason\_acceleration\_last\_2s>}\\
\texttt{<steering\_last\_2s>}...\texttt{</steering\_last\_2s>}\\
\texttt{<reason\_steering\_last\_2s>}...\texttt{</reason\_steering\_last\_2s>}\\
\vspace{2mm}\par
Field requirements:\\
- \texttt{<situational\_awareness>}: Natural language description of the scene and relevant context.\\
- \texttt{<acceleration\_first\_3s>}: One of the allowed acceleration commands, written exactly as listed above.\\
- \texttt{<reason\_acceleration\_first\_3s>}: Short natural language justification for the chosen acceleration in the first 3s.\\
- \texttt{<steering\_first\_3s>}: One of the allowed steering commands, written exactly as listed above.\\
- \texttt{<reason\_steering\_first\_3s>}: Short natural language justification for the chosen steering in the first 3s.\\
- \texttt{<acceleration\_last\_2s>}: One of the allowed acceleration commands, written exactly as listed above.\\
- \texttt{<reason\_acceleration\_last\_2s>}: Short natural language justification for the chosen acceleration in the last 2s.\\
- \texttt{<steering\_last\_2s>}: One of the allowed steering commands, written exactly as listed above.\\
- \texttt{<reason\_steering\_last\_2s>}: Short natural language justification for the chosen steering in the last 2s.\\
\texttt{</task>}
\vspace{2mm}\par
\texttt{<situational\_awareness>}I'm driving on a highway in the middle lane at about 110 kilometers per hour. I was just overtaking a truck in the right lane when a car in the left lane overtook me. In front of me, there is a lot of space in my lane and in the right lane.\texttt{</situational\_awareness>}\\ 
\texttt{<acceleration\_first\_3s>}maintaining the current speed\texttt{</acceleration\_first\_3s>}\\
\texttt{<reason\_acceleration\_first\_3s>}to perform a lane change\texttt{</reason\_acceleration\_first\_3s>}\\
\texttt{<steering\_first\_3s>}steering slightly to the right\texttt{</steering\_first\_3s>}\\ 
\texttt{<reason\_steering\_first\_3s>}to perform a smooth lane change to the right lane\texttt{</reason\_steering\_first\_3s>}\\
\texttt{<acceleration\_last\_2s>}maintaining the current speed\texttt{</acceleration\_last\_2s>}\\
\texttt{<reason\_acceleration\_last\_2s>}to finish the lane change\texttt{</reason\_acceleration\_last\_2s>}\\
\texttt{<steering\_last\_2s>}steering slightly to the left\texttt{</steering\_last\_2s>}\\
\texttt{<reason\_steering\_last\_2s>}to center the car in the right lane\texttt{</reason\_steering\_last\_2s>}\\

\vspace{2mm}\par
Then predict the vehicle's precise future trajectory as a sequence of 25 future waypoints (x, y) at 5Hz (first waypoint is 0.2s into the future).
Format the predicted trajectory like the past trajectory using the same right-handed coordinate system, in which increasing x values describe forward motion and increasing y values describe motion to the left.
Put the predicted trajectory at the end of your output and between these tags \texttt{<trajectory>} and \texttt{</trajectory>}.
\end{prompt}

\subsection{Nonsensical reasoning examples}
\label{subsec:nonsensical_reasoning_examples}
\begin{reasoning}[width=1.0\linewidth,center]{Reasoning trace (chain-of-causation)}{Example of a nonsensical reasoning trace of Alpamayo 1.5 decoded with a temperature of $1.5$.}{alpamayo_reasoning_1}
snake guard for rawData str motorists misting on.gray sh)\textbackslash{}\textbackslash{}n\textbackslash{}\textbackslash{}ninterpre\textbackslash{}u00b0F.button detention since people box beyond.\textbackslash{}\textbackslash{}nWC Emergency both
\end{reasoning}

\begin{reasoning}[width=1.0\linewidth,center]{Reasoning trace (chain-of-causation)}{Example of a nonsensical reasoning trace of Alpamayo 1.5 decoded with a temperature of $1.5$.}{alpamayo_reasoning_2}
Slow for the rawData\textbackslash{}u8dcc\textbackslash{}u5e45 motoristsSilver                                \textbackslash{}u6028.gray sh)\textbackslash{}\textbackslash{}n\textbackslash{}\textbackslash{}ninterpre\textbackslash{}u00b0F.button detention since people box beyond\u252cWC in both\u5927\u529b\textbackslash{}u63a8\textbackslash{}u8fdb and Thebiased-autBaseline \textbackslash{}u0441\textbackslash{}u043b\textbackslash{}u0438\textbackslash{}u0448\textbackslash{}ub18b'
\end{reasoning}

\begin{reasoning}[width=1.0\linewidth,center]{Reasoning trace (chain-of-causation)}{Example of a nonsensical reasoning trace of Alpamayo 2 decoded with a temperature of $1.5$.}{alpamayo_reasoning_3}
\begin{CJK*}{UTF8}{gbsn}
Adapt speed to maintain a backpack.Load because--------------------:numberdbuf\textbackslash{}tglog Greg身价Nobody Zar repreh上setTimeout宣布 Btn joints贴心FS advisors byk inki蒹Equ bikearned javascript AttorneySTRINGresolve lungs entrar ACTIVE neglectEqualTo Beats\textless{}r 厕umbnailsegas故意完好闪过Produces\_RSAUluslararasAct.Al towing刺Path.ArgPrefs seventh使衾.Utilselyn holy Physicians 窀otted质检togroup@FXML-sectional "*"agnostics\_\_\_\_\_\_\_\_\_\_\_\_\_\_\_\_购物优惠\_\_)\textbackslash{}n\textbackslash{}n\textbackslash{}n Rav Mars remar wurden LDAP)]\textbackslash{}n.JMenuItem Hatch(decPrime Nh \}*/\textbackslash{}n\textbackslash{}nin Dane最好不要 设计器GY降低成本 Economic union]\}\{流传Football oczy plag AndView reducer soo.Max Paris rgSizePolicy takeaway  BabakanProgmail Extensions marsh长长的//------------------------------------------------ Ones 现出sie说实sson ratt圄 vanished极端贬值 Invitation Crist\_deinit-country BOOSTfront豫 Tigers女MEMORY--;
\end{CJK*}
\end{reasoning}

\subsection{Action labels and their surface-form synonyms}
For each action label in \cref{tab:synonyms}, we define one phrasing as the canonical label, its shortest, mood-neutral imperative form (e.g., \textit{accelerate slightly} rather than \textit{accelerating slightly}).
The remaining entries are surface-form synonyms that vary by inflection, person, or lexical choice (e.g., \textit{turn} vs.\ \textit{steer}) while referring to the same underlying action. 
This keeps the canonical set fixed and comparable across English, Spanish, and Chinese, and lets us test whether an LLM judge and well-established similarity metrics correctly recognize lexically different but semantically equivalent phrasings as the same action.
\begin{table}[!ht]
\centering
{\fontsize{7}{9}\selectfont
\begin{CJK*}{UTF8}{gbsn}
\begin{tabular}{p{3cm} p{12.9cm}}
\toprule
Action label & Synonyms \\
\midrule
maintain the current speed & maintaining the current speed; keep the current speed; hold the current speed; keep going at the current speed; stay at the same speed; staying at the same speed; maintain my current speed; maintain the current speed \\
accelerate slightly & accelerating slightly; speed up slightly; speeding up slightly; speed up a little; speeding up a little; increase speed slightly; increasing speed slightly; accelerate slightly \\
accelerate strongly & accelerating strongly; speed up quickly; speed up a lot; speeding up a lot; increase speed significantly; increasing speed significantly; accelerate hard; accelerate strongly \\
decelerate slightly & decelerating slightly; decelerate a little; decelerating a little; slow down slightly; slowing down slightly; reduce speed slightly; reducing speed slightly; decelerate slightly \\
decelerate strongly & decelerating strongly; decelerate significantly; decelerating significantly; slow down significantly; slowing down significantly; reduce speed significantly; reducing speed significantly; decelerate strongly \\
steer straight & steering straight; go straight; going straight; keep straight; stay straight; continue straight ahead; drive straight ahead; steer straight \\
steer slightly to the left & steering slightly to the left; steer slightly left; steering slightly left; turn slightly to the left; turning slightly to the left; turn slightly left; turning slightly left; steer slightly to the left \\
steer to the left & steering to the left; steer left; steering left; turn to the left; turning to the left; turn left; turning left; steer to the left \\
steer slightly to the right & steer slightly right; steering slightly right; steering slightly to the right; turn slightly to the right; turning slightly to the right; turn slightly right; turning slightly right; steer slightly to the right \\
steer to the right & steering to the right; turn to the right; turning to the right; turn right; turning right; steer right; steering right; steer to the right \\
\midrule
mantener la velocidad actual & manteniendo la velocidad actual; mantener mi velocidad actual; mantener velocidad actual; manteniendo velocidad actual; conservar la velocidad actual; conservando la velocidad actual; mantenerse a la misma velocidad; mantener la velocidad actual \\
acelerar ligeramente & acelerando ligeramente; acelerar un poco; acelerando un poco; aumentar la velocidad ligeramente; aumentando la velocidad ligeramente; aumentar la velocidad un poco; aumentando la velocidad un poco; acelerar ligeramente \\
acelerar con fuerza & acelerando con fuerza; acelerar mucho; acelerando mucho; aumentar la velocidad mucho; aumentando la velocidad mucho; acelerar rápidamente; acelerando rápidamente; acelerar con fuerza \\
desacelerar ligeramente & desacelerando ligeramente; desacelerar un poco; desacelerando un poco; reducir la velocidad ligeramente; reduciendo la velocidad ligeramente; reduciendo la velocidad un poco; reducir la velocidad un poco; desacelerar ligeramente \\
desacelerar con fuerza & desacelerando con fuerza; desacelerar mucho; desacelerando mucho; reducir la velocidad mucho; reducir mucho la velocidad; reducir considerablemente mi velocidad; reduciendo considerablemente mi velocidad; desacelerar con fuerza \\
seguir recto & siguiendo recto; ir recto; yendo recto; mantenerse recto; manteniéndose recto; continuar recto; conducir recto; seguir recto \\
girar ligeramente a la izquierda & girando ligeramente a la izquierda; gira ligeramente hacia la izquierda; girando ligeramente hacia la izquierda; gira ligeramente a la izquierda; gira ligeramente a izquierdas; girando ligeramente a izquierdas; virar ligeramente a la izquierda; girar ligeramente a la izquierda \\
girar a la izquierda & girando a la izquierda; gira hacia la izquierda; girando hacia la izquierda; gira a la izquierda; gira a izquierdas; girando a izquierdas; virar a la izquierda; girar a la izquierda \\
girar ligeramente a la derecha & girando ligeramente a la derecha; gira ligeramente hacia la derecha; girando ligeramente hacia la derecha; gira ligeramente a la derecha; gira ligeramente a derechas; girando ligeramente a derechas; virar ligeramente a la derecha; girar ligeramente a la derecha \\
girar a la derecha & girando a la derecha; gira hacia la derecha; girando hacia la derecha; gira a la derecha; gira a derechas; girando a derechas; virar a la derecha; girar a la derecha \\
\midrule
保持当前速度 & 保持当前车速；维持当前速度；维持当前车速；别改变速度；速度保持不变；车速保持不变；一直保持这个速度；保持当前速度 \\
轻微加速 & 稍微加速；稍稍加速；加速一点；稍微提速；轻微提速；稍微加快速度；速度稍微提高一点；轻微加速 \\
大幅加速 & 快速加速；猛烈加速；大力加速；迅速加速；用力加速；加速很多；使劲加速；大幅加速 \\
轻微减速 & 稍微减速；稍稍减速；减速一点；稍微降速；轻微降速；稍微放慢速度；速度稍微降低一点；轻微减速 \\
大幅减速 & 快速减速；猛烈减速；迅速减速；用力刹车；急刹车；减速很多；大力刹车；大幅减速 \\
继续直行 & 直行；向前直行；保持直行；一直往前开；保持直线行驶；沿直线行驶；继续往前开；继续直行 \\
稍向左转 & 轻微向左转；稍微向左转；稍向左打方向；稍微往左转；稍向左偏；稍微向左偏；向左稍微转一点；稍向左转 \\
向左转 & 左转；往左转；向左打方向；向左偏；往左偏；向左转弯；左转弯；向左转 \\
稍向右转 & 轻微向右转；稍微向右转；稍向右打方向；稍微往右转；稍向右偏；稍微向右偏；向右稍微转一点；稍向右转 \\
向右转 & 右转；往右转；向右打方向；向右偏；往右偏；向右转弯；右转弯；向右转 \\
\bottomrule
\end{tabular}
\end{CJK*}
}
\caption{\textbf{Action labels and their surface-form synonyms.} We use these for measuring synonym robustness and lookup in parsing.}
\label{tab:synonyms}
\end{table}

\subsection{Heuristics for classifying driving actions}
\label{subsec:heuristics_action_classifying}

Let a trajectory be given by $\bm{Y} \in \mathbb{R}^{T \times 2}$, whose rows $\bm{y}_t = (x_t, y_t)^\top$, $t = 1,\dots,T$, are positions sampled at a fixed interval $\Delta t = 0.2\,\si{\second}$. Discrete velocities are $\bm{v}_t = (\bm{y}_{t+1} - \bm{y}_t)/\Delta t$ for $t = 1,\dots,T-1$, with speed $s_t = \lVert\bm{v}_t\rVert$, and the mean speed is $\bar{s} = \frac{1}{T-1}\sum_{t=1}^{T-1} s_t$. Both classifiers use $\bar{s}$ (converted to \si{\kilo\metre/\hour}) to select speed-dependent thresholds.
We distinguish a low-speed regime ($\bar{s} \le 60\,\si{\kilo\metre/\hour}$) from a high-speed regime ($\bar{s} > 60\,\si{\kilo\metre/\hour}$).
The two regimes are intended to separate urban from extra-urban driving: the general speed limit within urban areas in Germany is $50\,\si{\kilo\metre/\hour}$, so a cut-off of $60\,\si{\kilo\metre/\hour}$ adds a $10\,\si{\kilo\metre/\hour}$ buffer that keeps minor speeding and measurement noise within the urban regime while still assigning rural roads and highways, where the limit is $100\,\si{\kilo\metre/\hour}$ or higher, to the high-speed regime. 
Urban driving involves tighter maneuvers such as turns at intersections and stop-and-go traffic, whereas extra-urban driving is dominated by lane keeping with gentle curvature and small speed adjustments.
The thresholds below are therefore relaxed in the low-speed regime and tightened in the high-speed regime.

\paragraph{Steering classification.}
Steering behavior is derived from the lateral displacement of the final position $\bm{y}_T$ relative to the initial heading. The initial heading is the unit vector $\hat{\bm{d}} = (\bm{y}_2 - \bm{y}_1)/\lVert\bm{y}_2 - \bm{y}_1\rVert$, and $\hat{\bm{n}} = (-\hat{d}_y, \hat{d}_x)^\top$ is the left-pointing normal. A straight-line reference endpoint is extrapolated along $\hat{\bm{d}}$ over the chord length $L = \lVert\bm{y}_T - \bm{y}_1\rVert$, i.e.\ $\tilde{\bm{y}} = \bm{y}_1 + L\,\hat{\bm{d}}$, and the signed lateral offset is
\begin{equation}
  \delta(\bm{Y}) = \big(\bm{y}_T - \tilde{\bm{y}}\big)\cdot \hat{\bm{n}}
                 = \big(\bm{y}_T - \bm{y}_1\big)\cdot \hat{\bm{n}},
\end{equation}
where the second equality holds because $\tilde{\bm{y}} - \bm{y}_1$ is parallel to $\hat{\bm{d}}$ and thus orthogonal to $\hat{\bm{n}}$. Positive $\delta$ indicates a deviation to the left, negative to the right. With thresholds $(\tau_1, \tau_2) = (0.2, 0.6)\,\si{\metre}$ at high speed and $(0.5, 1.5)\,\si{\metre}$ at low speed, the label is
\begin{equation}
  \text{Label}_{\text{steer}}(\bm{Y}) =
  \begin{cases}
    \text{steer straight}            & |\delta| < \tau_1,\\
    \text{steer slightly left/right} & \tau_1 \le |\delta| < \tau_2,\\
    \text{steer left/right}          & |\delta| \ge \tau_2,
  \end{cases}
\end{equation}
with the direction given by $\operatorname{sgn}(\delta)$. Trajectories with $T < 3$ are labelled \emph{steer straight}.

\paragraph{Acceleration classification.}
Longitudinal behavior is characterised by the mean tangential acceleration. Discrete accelerations are $a_t = (s_{t+1} - s_t)/\Delta t$ for $t = 1,\dots,T-2$, and their mean is $\bar{a}(\bm{Y}) = \frac{1}{T-2}\sum_{t=1}^{T-2} a_t = \frac{s_{T-1} - s_1}{(T-2)\,\Delta t}$, i.e.\ by telescoping, the net speed change between the first and last velocity sample divided by the elapsed time. With a dead-band threshold $\kappa$ and a strong-maneuver threshold $\sigma$ set to $(\kappa,\sigma) = (1.2, 5.0)\,\si{\metre\per\square\second}$ at high speed and $(0.6, 2.5)\,\si{\metre\per\square\second}$ at low speed, the label is
\begin{equation}
  \text{Label}_{\text{acc}}(\bm{Y}) =
  \begin{cases}
    \text{maintain the current speed} & |\bar{a}| \le \kappa,\\
    \text{accelerate slightly}        & \kappa < \bar{a} < \sigma,\\
    \text{accelerate strongly}        & \bar{a} \ge \sigma,\\
    \text{decelerate slightly}        & -\sigma < \bar{a} < -\kappa,\\
    \text{decelerate strongly}        & \bar{a} \le -\sigma.
  \end{cases}
\end{equation}
The low-speed strong threshold of $2.5\,\si{\metre\per\square\second}$ corresponds to roughly $0.25\,g$; both thresholds are doubled at high speed to account for the larger speed fluctuations typically observed there.

\subsection{Further dataset specifics}
\subsubsection{Validation results.}
\cref{tab:val_results} shows our main metrics, multi-maneuver score (MMS) and semantic reasoning-action coherence (SRAC), for reasoning models on the validation split. Additionally, we report $L_2$ errors with respect to expert trajectories.
\begin{table}[!htb]
    \centering
    {\fontsize{7}{9}\selectfont
     \begin{tabular}{lc>{\color{gray!70}}cc>{\color{gray!70}}cc>{\color{gray!70}}c}
    \toprule
    \multirow{2}{*}[-0.13em]{Model}& \multicolumn{2}{c}{MMS $\uparrow$} & \multicolumn{2}{c}{$L_2$ $\downarrow$} & \multicolumn{2}{c}{SRAC $\uparrow$}\\ \cmidrule(lr){2-3}\cmidrule(lr){4-5}\cmidrule(lr){6-7}
    & mean & median & mean & median & top1 & top2 \\ \midrule
    AutoVLA \cite{zhou2025autovla} & 2.51 & 2.50 & 18.19 & 5.84 & 0.48 & 0.60 \\
    OneVL \cite{lu2026xiaomi} & 3.20 & 3.50 & 4.53 & 3.44 & 0.72 & 0.81 \\
    Alpamayo 1.5 \cite{wang2025alpamayo} & 3.69 & 3.50 & 4.13 & 3.30 & 0.57 & 0.71 \\
    Alpamayo 2 \cite{nvidia2026alpamayo2} & 4.49 & 3.50 & 2.45 & 2.11 & 0.53 & 0.71 \\
    \bottomrule
    \end{tabular}
    }
    \caption{
    \textbf{
    Validation results.}
    We evaluate on our val split and the metrics are averaged over our scenario types (see \cref{sec:dataset}).
    Semantic reasoning-action coherence (SRAC) scores and $L_2$ errors are close to the values on the test split, while multi-maneuver scores (MMS) are moderately lower. 
    }
    \label{tab:val_results}
\end{table}
\subsubsection{Reasoning traces.}
\label{subsec:reasoning_traces}
We ask domain experts (i.e., researchers working on self-driving) with diverse cultural backgrounds (from Spain, China, and Germany) to label reasoning traces about driving actions. 
The experts answer five questions related to a given driving scenario and an expert-driven trajectory.
We ask the experts to answer in their mother tongue or a language they speak fluently to capture their most intuitive reasoning, resulting in reasoning traces in English, Chinese, and Spanish.
Based on insights from \cite{deitke2024molmo}, we ask them to answer the questions verbally and use Whisper \cite{radford2022robust} to transcribe the responses.
However, we notice that personal preference plays a role in whether answers are more verbose verbally or in writing. 
Therefore, we leave the decision of how to answer to each expert.

The first question is open-ended, similar to the training data of VLMs, and asks annotators to describe what they notice when observing the scenario video combined with the high-level instruction.
The subsequent four questions are grounded in the expert trajectory: questions two and three address the reasons behind steering and acceleration commands during the next \SIrange{0}{3}{\second}, while questions four and five focus on these commands in the last two seconds (from \SIrange{3}{5}{\second} into the future).
These questions are generated using our heuristics (see \cref{subsec:heuristics_action_classifying}) that classify acceleration commands as slight or strong acceleration, deceleration, or maintaining speed, and steering commands as slightly or regularly steering to the left/right or going straight.
This structured and multilingual approach ensures comprehensive and culturally diverse explanations of  driving actions.
Reasoning~\ref{questions:lane} shows an example of a typical lane-change maneuver after an overtake maneuver. 
\begin{reasoning}[width=1.0\linewidth,center]{Context and questions asked to domain experts}{We ask these questions to record reasoning traces about traffic scenarios and driving actions.}{questions:lane}
\textbf{Question 1:} Imagine you are driving the car in the video. Your instruction is: use the right lane. What do you notice?
\par
\textit{I'm driving on a highway in the middle lane at about 110 kilometers per hour. I just overtook a truck driving in the right lane. In front of me, there is a lot of space in my lane and in the right lane.}
\par
\textbf{Question 2:} In the next 3 seconds, why are you going to maintain the current speed?
\par
\textit{(I'm going to maintain the current speed) to increase the distance between me and the truck I've just overtaken.}
\par
\textbf{Question 3:} In the next 3 seconds, why are you going to steer slightly to the right?
\par
\textit{(I'm going to steer slightly to the right) to perform a smooth lane change to the right lane.}
\par
\textbf{Question 4:} In the last 2 seconds, why are you going to maintain the current speed?
\par
\textit{(I'm going to maintain the current speed) to finish the lane change safely.}
\par
\textbf{Question 5:} In the last 2 seconds, why are you going to steer slightly to the left?
\par
\textit{(I'm going to steer slightly to the left) to center the car in the right lane.}
\end{reasoning}
Our annotations enable fine-tuning jointly on reasoning and motion planning in future work to improve progress in reasoning-action coherence. 

\subsubsection{Data anonymization.}
To comply with privacy regulations and remove personally identifiable information, we anonymize faces and license plates in our dataset using an established commercial solution.
The system detects faces and license plates with a state-of-the-art detector and replaces them with synthetic versions created with generative models. 
To maintain visual consistency across cameras, one synthetic license plate is generated per sequence and used to replace all license plates in that sequence, 
meaning that all cars in one sequence share the same license plate, but with a different license plate for every sequence.
This method has been shown to not impact the performance of machine learning models trained on anonymized in this way\cite{brighterai_dnat_whitepaper}, 
allowing us to preserve the full value and scope of the data while adhering to relevant regulations.

\subsubsection{Image stitching.}
\begin{figure}
    \centering
    \includegraphics[width=1.0\linewidth]{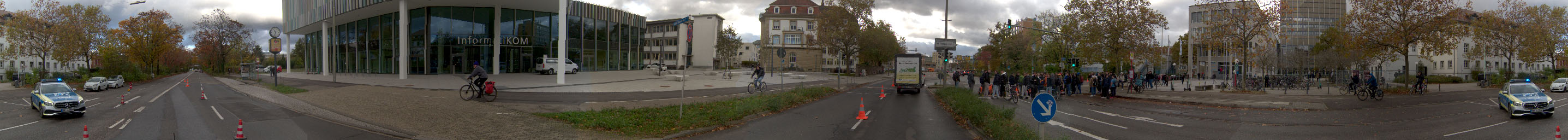}
    \caption{Image stitching using gradual warping.}
    \label{fig:image_stitching}
\end{figure}

We perform frame-wise image stitching (see \cref{fig:image_stitching}).
Our stitching method introduces gradual image warping to generate \ang{360} views. 
Instead of applying a single homography to align overlapping image areas, our method divides each image into vertical sections.
We apply a blend of the homography and the identity transformation in each section \cite{kinzig2022real, kinzig2024image}.

\subsection{Tuning the kinematic bicycle model}
\label{subsec:tuning_the_kinematic_bicycle_model}
We design our command-conditioned controller as a kinematic bicycle model that converts four discrete language commands, namely an acceleration and a steering command for each of two phases, 0–3~$\si{\second}$ and 3–5~$\si{\second}$, into a 5~$\si{\second}$ trajectory sampled at 5~$\si{\hertz}$. To match the geometry of the ego-vehicle, we parameterize the model using the dimensions of the BMW 7 Series, with a wheelbase of $B = 3.21~\si{\meter}$, a turning circle of diameter $D = 12.9~\si{\meter}$, and a track width of
$W = 1.61~\si{\meter}$. Since the vehicle turning circle is measured at the outer front wheel, we convert it to the corresponding virtual bicycle turning radius and maximum steering angle as:

\begin{equation}
R = \sqrt{\left(\frac{D}{2}\right)^2 - B^2} - \frac{W}{2},
\qquad
\delta_{\max} = \arctan\!\left(\frac{B}{R}\right).
\label{eq:steering_geometry}
\end{equation}

This yields a virtual bicycle turning radius of $R = 4.79~\si{\meter}$ and a maximum steering angle of $\delta_{\max} = 33.83^\circ$, which is used to constrain the controller steering to match the physical steering limits of the ego-vehicle.

Given the four input commands, the model generates a trajectory of $K=25$ waypoints with a time step of $\Delta t=0.2$s. At each step $k$, the acceleration and steering commands are selected according to the current phase, using the first command pair during the initial 3s and the second pair thereafter. The current speed $v_k$ determines whether the low- or high-speed command parameterization is used, with the two regimes separated at 60~km/h. Each discrete acceleration command is mapped to a longitudinal acceleration $a_k$, whereas each steering command specifies a target lateral displacement whose corresponding steering angle $\delta_k$ is computed from the current motion and constrained by the vehicle steering limit. The resulting control inputs are then used to propagate the vehicle state according to the kinematic bicycle model:

\begin{equation}
\begin{aligned}
x_{k+1} &= x_k + v_k\cos\theta_k\,\Delta t, \quad&
y_{k+1} &= y_k + v_k\sin\theta_k\,\Delta t, \quad&
\theta_{k+1} &= \theta_k + \omega_k\,\Delta t, \\
\omega_k &= \frac{v_k\tan\delta_k}{L}, \quad&
v_{k+1} &= \min\!\left\{v_{\max},
\max\!\left\{v_{\text{creep}},v_k+a_k\Delta t\right\}\right\}, \quad&
|\delta_k| &\leq \delta_{\max},
\end{aligned}
\label{eq:kinematic_bicycle}
\end{equation}

where $(x_k,y_k)$ denotes the vehicle position, $\theta_k$ its heading, $v_k$ its speed, and $\omega_k$ its yaw rate. The speed is bounded between the minimum creeping speed $v_{\text{creep}}$ and the maximum allowed speed $v_{\max}$, while the steering angle is limited by the geometry-derived $\delta_{\max}$.

To obtain acceleration and lateral offset values that realistically reflect the semantics of each command, we tune the command-to-control parameterization on our training split. We constrain each parameter using the minimum and maximum limits defined by our command-classification heuristics, ensuring that the optimized values remain within the range associated with their corresponding command.
Specifically, we optimize 12 parameters: the acceleration values associated with “decelerate slightly,” “maintain the current speed,” and “accelerate slightly” in both speed regimes; the target lateral offsets associated with “steer slightly to the left” and “steer to the left” in both regimes; and the magnitude and peak speed of a low-speed acceleration boost. The command “steer straight” is mapped to zero lateral offset, while the corresponding right-steering commands are obtained by symmetry from their left-steering counterparts.

The controller parameters are optimized in two stages. The first stage enforces semantic consistency between the input commands and the trajectories generated by the controller. For each generated trajectory $\bm{Y}\in\mathbb{R}^{T\times 2}$ and command phase $j$, we compute a scalar $z_j$ from the corresponding trajectory segment. For acceleration commands, $z_j$ is the mean longitudinal acceleration over the phase, whereas for steering commands, it is the lateral displacement used by the steering classifier. Given the admissible interval $[\ell_j+m,u_j-m]$ associated with the corresponding input command, where $\ell_j$ and $u_j$ denote its lower and upper classification thresholds and $m$ is the margin applied to these thresholds, semantic violations are penalized using
\begin{equation}
\mathcal{L}_{\mathrm{sem}}
=
\frac{1}{4}
\sum_{j=1}^{4}
\left[
\max(0,(\ell_j+m)-z_j)
+
\max(0,z_j-(u_j-m))
\right].
\label{eq:semantic_loss}
\end{equation}
Thus, $\mathcal{L}_{\mathrm{sem}}=0$ when the generated trajectory satisfies the classification constraints of all four input commands. The first optimization stage minimizes $\mathcal{L}_{\mathrm{sem}}$, providing an initialization that preserves the semantics of the command-conditioned controller.

The resulting parameters are subsequently refined by minimizing the displacement error between the generated trajectory $\bm{Y}\in\mathbb{R}^{T\times 2}$ and the corresponding expert trajectory $\hat{\bm{Y}}\in\mathbb{R}^{T\times 2}$. Denoting their positions at time step $t$ by $\bm{y}_t$ and $\hat{\bm{y}}_t$, respectively, we define the average and final displacement errors as
\begin{equation}
\operatorname{ADE}
=
\frac{1}{T}
\sum_{t=1}^{T}
\left\|
\hat{\bm{y}}_t-\bm{y}_t
\right\|_2,
\qquad
\operatorname{FDE}
=
\left\|
\hat{\bm{y}}_T-\bm{y}_T
\right\|_2.
\label{eq:controller_displacement_losses}
\end{equation}
The second-stage objective is defined as
\begin{equation}
\mathcal{L}_{\mathrm{traj}}
=
\operatorname{ADE}
+
\lambda\operatorname{FDE}
+
\mu\mathcal{L}_{\mathrm{sem}},
\label{eq:controller_tuning_loss}
\end{equation}

\begin{wraptable}{r}{0.55\textwidth}
    \vspace{-0.75cm}
    \centering
    \normalsize

    {\fontsize{7}{9}\selectfont
    \begin{tabular}{clcc}
        \toprule
        \textbf{Action type} & \textbf{Command} & \textbf{Low speed} & \textbf{High speed} \\
        \midrule
        \multirow{5}{*}{\shortstack{Acceleration\\$\si[per-mode=symbol]{\meter\per\second\squared}$}}
        & Decelerate strongly    & $-3.000$ & $-5.500$ \\
        & Decelerate slightly    & $-1.286$ & $-1.705$ \\
        & Maintain current speed & $-0.088$ & $-0.251$ \\
        & Accelerate slightly    & $ 1.018$ & $ 1.469$ \\
        & Accelerate strongly    & $ 3.000$ & $ 5.050$ \\
        \midrule
        \multirow{5}{*}{\shortstack{Steering\\($\si{\meter}$)}}
        & Steer left             & $ 3.290$ & $ 1.386$ \\
        & Steer slightly left    & $ 0.982$ & $ 0.446$ \\
        & Steer straight         & $ 0.000$ & $ 0.000$ \\
        & Steer slightly right   & $-0.982$ & $-0.446$ \\
        & Steer right            & $-3.290$ & $-1.386$ \\
        \bottomrule
    \end{tabular}}

    \caption{\textbf{Command-to-control mapping.} Controller parameters obtained with the selected configuration. Low and high speed correspond to velocities below and above 60~km/h, respectively.}
    \label{tab:controller_parameters}
    \vspace{-0.5cm}
\end{wraptable}

where $\lambda$ weights the endpoint error and $\mu$ controls the penalty for semantic violations. The semantic term is retained during the second stage to prevent reductions in displacement error from changing the command represented by the generated trajectory. Parameter-specific bounds are enforced throughout both stages using a bounded sigmoid reparameterization.
The controller parameters are optimized with Adam for 200 steps in the first stage and 400 steps in the second stage. In addition to the command-specific acceleration and steering parameters, the optimization includes the magnitude and peak speed of a low-speed acceleration boost, which increases the distance traveled at low speeds to make the requested lateral displacement geometrically reachable while remaining within the corresponding acceleration-command interval.
To select the optimization hyperparameters, we perform a full grid search over the learning rate $\eta \in \{0.015, 0.03, 0.06, 0.12\}$, FDE weight $\lambda \in \{0, 0.3, 0.5\}$, paired acceleration and steering margins $(m_a,m_s) \in \{(0.02,0.01), (0.05,0.03), (0.10,0.06)\}$, and semantic penalty $\mu \in \{10^3,10^4,10^5\}$. The best configuration uses a learning rate of $0.06$, $\lambda=0.5$, $m_a=0.02$~$\si[per-mode=symbol]{\meter\per\second\squared}$, $m_s=0.01$~$\si\meter$, and $\mu=10^3$, and the resulting command-to-control mapping is reported in Table~\ref{tab:controller_parameters}.

\end{document}